%% file: main.tex
\documentclass[aps,pra,twocolumn,amsmath,amssymb,superscriptaddress,reprint,longbibliography]{revtex4-2}

\usepackage[utf8]{inputenc}
\usepackage{graphicx}
\usepackage[colorlinks=true,citecolor=blue]{hyperref}
\usepackage{subcaption}
\usepackage{lipsum}
\usepackage{dblfloatfix}
\usepackage{float}

\begin{document}

\title{Computational lexical analysis of Flamenco genres}

\author{Pablo Rosillo-Rodes}
\email{prosillo@ifisc.uib-csic.es}
\author{Maxi San Miguel}
\author{David Sánchez}

\affiliation{
 Institute for Cross-Disciplinary Physics and Complex Systems IFISC (UIB-CSIC), Campus Universitat de les Illes Balears, E-07122 Palma de Mallorca, Spain
}
\date{\today}

\begin{abstract}
Flamenco, recognized by UNESCO as part of the Intangible Cultural Heritage of Humanity, is a profound expression of cultural identity rooted in Andalusia, Spain. However, there is a lack of quantitative studies that help identify characteristic patterns in this long-lived music tradition. In this work, we present a computational analysis of Flamenco lyrics, employing natural language processing and machine learning to categorize over 2000 lyrics into their respective Flamenco genres, termed as \textit{palos}. Using a Multinomial Naive Bayes classifier, we find that lexical variation across styles enables to accurately identify distinct \textit{palos}. More importantly, from an automatic method of word usage, we obtain the semantic fields that characterize each style. Further, applying a metric that quantifies the inter-genre distance we perform a network analysis that sheds light on the relationship between Flamenco styles. Remarkably, our results suggest historical connections and \textit{palo} evolutions. Overall, our work illuminates the intricate relationships and cultural significance embedded within Flamenco lyrics, complementing previous qualitative discussions with quantitative analyses and sparking new discussions on the origin and development of traditional music genres.
\end{abstract}

\maketitle

\section{Introduction}
\label{sec:intro}

In the past decade, there has been a significant drive to utilize computational techniques to investigate intangible cultural heritage. These efforts encompass a broad range of activities, from the development of cultural platforms and databases~\cite{Serra2014, Kroher2016, Abdallah2017, Oramas2018, Sarti2022, Eero2023} to the application of social signal processing to oral history collections~\cite{Pessanha2021}. Additionally, there have been explorations into artistic folklore studies, such as various approaches for analyzing dance movements~\cite{Aristidou2015, Stepputat2019, Parthasarathy2023}, analysis of musical metrics of contemporary western popular music through the decades~\cite{Serra2012, Buongiorno2022}, and advancements in music innovation studies~\cite{Sobchuk2022}. Our research aims to contribute to the understanding of the lyrics of Flamenco, a musical tradition deeply rooted in culture but scarcely studied with quantitative methods. 

Flamenco, a rich oral musical tradition originating in Andalusia, a region in southern Spain, reflects a diverse tapestry of cultural influences. Throughout the centuries, this region, along with its musical expression, has been significantly shaped by various cultural settlements. Notably, the influences of Jews, Muslims, Christians, and folkloric music from different Spanish regions are discernible contributors, yet undeniably, the cultural imprint of Andalusian gipsies is deeply embedded in the essence of Flamenco~(\cite{Gamboa2005},~p.~469-473). Recognizing its distinct features and profound significance for the cultural identity of Andalusia, Flamenco was honored with inclusion in the UNESCO List of Intangible Cultural Heritage of Humanity in 2010~\cite{UNESCOFlamenco}.

This intricate musical form unfolds across a wide spectrum of settings, ranging from lively \textit{fiestas} (private parties) and intimate \textit{tablaos} (Flamenco venues) to expansive concerts and elaborate productions in theaters. In each of these diverse contexts, the fundamental components of Flamenco music emerge: the \textit{cante} (singing), the \textit{toque} (instrument playing), and the \textit{baile} (dance). Here, we focus on the song lyrics, and we demonstrate how they are intimately related to the different Flamenco genres. 

Flamenco, while inherently individualistic, adheres to a highly structured framework. The tradition, characterized by a complex organization of styles and structures, allows for extensive improvisation within defined parameters. This implicit knowledge forms the foundation for spontaneous artistic expressions, wherein practitioners skillfully blend the fixed rhythmic, melodic, and harmonic structures of a specific style with a personalized set of expressive resources. This is naturally reflected in the lyrics, e.g., in formulaic expressions rooted in the traditional genre structure as we will see below.

In the Flamenco jargon genres or styles are called \textit{palos}. Criteria adopted to define \textit{palos} are rhythmic patterns, chord progressions, lyrics and their poetic structure, and geographical origin~\cite{Oramas2018}. Due to its oral and traditional transmission and the inherent personal style of interpretation of the singing, genre classification of Flamenco music is often subject to personal interpretation and previous experience of the listener in Flamenco gatherings, leading even to academic discussions about the proper classification of some \textit{palos}~\cite{Fernandez2015}. Here, following a quantitative approach, we investigate the Flamenco genre classification. Importantly, we find that Flamenco styles are characterized by unique lexical features as shown in their lyrics.

The foundational musical concepts and the evolutionary trajectory of Flamenco music remain largely uncharted. This dearth of documented information poses significant constraints on traditional musicological studies within the subject while serving as a compelling impetus for the development of computational tools dedicated to the description and analysis of Flamenco. While computational tools devoted to analyzing Flamenco art cover a wide range of specialties, such as biomechanical studies of \textit{baile} (dance)~\cite{Forczek2021}, analysis of rhythmic similarity of \textit{cantes} (singing)~\cite{Diaz2004, Guastavino2009}, and creation of musical and informational databases~\cite{Kroher2016, Oramas2018}, to the best of our knowledge there is inexistent research addressing the computational analysis of Flamenco lyrics. This is the gap that we want to fill with our work.

Flamenco, as an oral music tradition, involves the transmission of songs and musical nuances across generations. Consequently, written scores are a rarity, given the intricate and complex melodic ornamentation inherent in vocal melodies. Even the records of renowned Flamenco festivals lack information about some features of the performances, including the lyrics~\cite{Gimenez2020}. The process of manual annotation, essential for capturing these subtleties, is not only exceedingly time-consuming but also inevitably involves a subjective layer of interpretation.

In virtue of their aforementioned oral transmission which, while allowing for improvisation, firmly respects tradition, Flamenco lyrics are proven to describe the sociocultural context of their creators~\cite{Polackova2011, Homann2020}. Our interest lies in the systematic and large-scale analysis of Flamenco lyrics, to help us understand the different cultures and identities for which they constitute a historical vehicle of transmission. In this work, we quantitatively analyze the features of a corpus consisting of over 2000 manually-transcribed Flamenco lyrics with extensive metadata (title, album, singer, \textit{palo}, etc.) available online~\cite{corpus}.

To analyze large volumes of text, we use Natural Language Processing (NLP) tools. The newborn field of Lyrics Information Processing (LIP)~\cite{Watanabe2020} has proven to be successful in retrieving all kinds of information through the use of NLP for a computational analysis of lyrics outside Flamenco. Related works on automatic analysis of lyrics cover a wide range of applications such as the author classification of German songs~\cite{Mendhakar2023}, automatic detection of sexism in English songs~\cite{Betti2023}, the age classification of Indonesian lyrics~\cite{Thirafi2018} or the identification of different writing styles in Taylor Swift's songs~\cite{Kendong2023}, among others~\cite{Mahedero2005, Fell2014, Ahmed2022, Parada2024}. However, to our knowledge, the computational analysis of Flamenco singers' production is limited to studies involving analysis of melody and related features~\cite{Gomez2011, Mora2016} rather than the content of their lyrics. We thus face an absence of LIP applications for Flamenco lyrics analysis. 

Given the situation described above, our Research Questions (RQ) are:
\begin{enumerate}
    \item Can Flamenco songs be properly assigned to their corresponding \textit{palo} solely by means of their lexicon? If so, what is the characteristic lexicon of each \textit{palo}?
    \item What relationships exist between the main Flamenco \textit{palos} based on their lexicon?
\end{enumerate}

To answer RQs 1 and 2 we analyze the corpus~\cite{corpus} and train a Multinomial Naive Bayes algorithm \cite{Kibriya2005}, a machine learning classification model widely used in text analysis. 

We demonstrate that it is possible to differentiate between the main Flamenco genres and to properly classify them solely by their lyrics, and we determine the characteristic lexicon that allows us to do so (RQ1). We also compute lexical distances and propose a matrix of \textit{palo} inter-relationships. These measures allow us to quantitatively prove some well-established facts about the origin and kinship of certain Flamenco genres (RQ2) and also propose open questions about some others.

\section{Results}
\label{sec:results}

In this section, we show the results obtained in our analysis of Flamenco lyrics to answer RQs 1 and 2. First, we explore the vocabulary richness and distribution of Flamenco genres through a computational analysis of our lyrics dataset. Later, we use a machine learning model to classify Flamenco songs based solely on their lexicon. Finally, we examine the relationships between the different \textit{palos} based on their characteristic vocabulary.

\subsection{Dataset}

We analyze a corpus of Flamenco lyrics curated by Norman P. Kliman on his website~\cite{corpus}. This corpus contains 3380 lyrics spanning 78 different styles, or \textit{palos}. The lyrics were transcribed by both dedicated aficionados and Kliman himself. Metadata such as author, album, year, and \textit{palo} was not assigned by the transcribers, but rather reflects the information provided by the original record label or other publishing sources. As mentioned in Sec.~\ref{sec:intro}, \textit{palo} cathegorization is subject of melodical and musical interpretations, which can vary depending on the listener and the tradition. Importantly, we use the \textit{palo} metadata provided by the record labels of the albums as the ground truth for the supervised learning algorithm (see Sec.~\ref{sec:results_paloclass} and Appendix~\ref{app:mnb}).  The corpus features 276822 tokens (words) and 12981 types (unique words). The corpus is varied and diverse, exhibiting a structure characteristic of natural language corpora (see Appendix~\ref{sec:emphapax}). 

As we can see in Fig.~\ref{fig:palohist}, not all the 78 different \textit{palos} are equally represented, and only 8 of them count on more than 100 lyrics. Due to this, we decided to perform our analysis on the most represented genres, i.e., \textit{bulerías}, \textit{soleá}, \textit{seguiriyas}, \textit{fandangos}, \textit{tangos}, \textit{tientos}, \textit{malagueñas}, and \textit{alegrías}. This set of the 8 most represented \textit{palos} constitutes the reduced corpus to which, from now on, we will refer as \textit{the corpus} for the sake of brevity.

\begin{figure}[t]
  \centering
  \includegraphics[height=180pt, angle=-90]{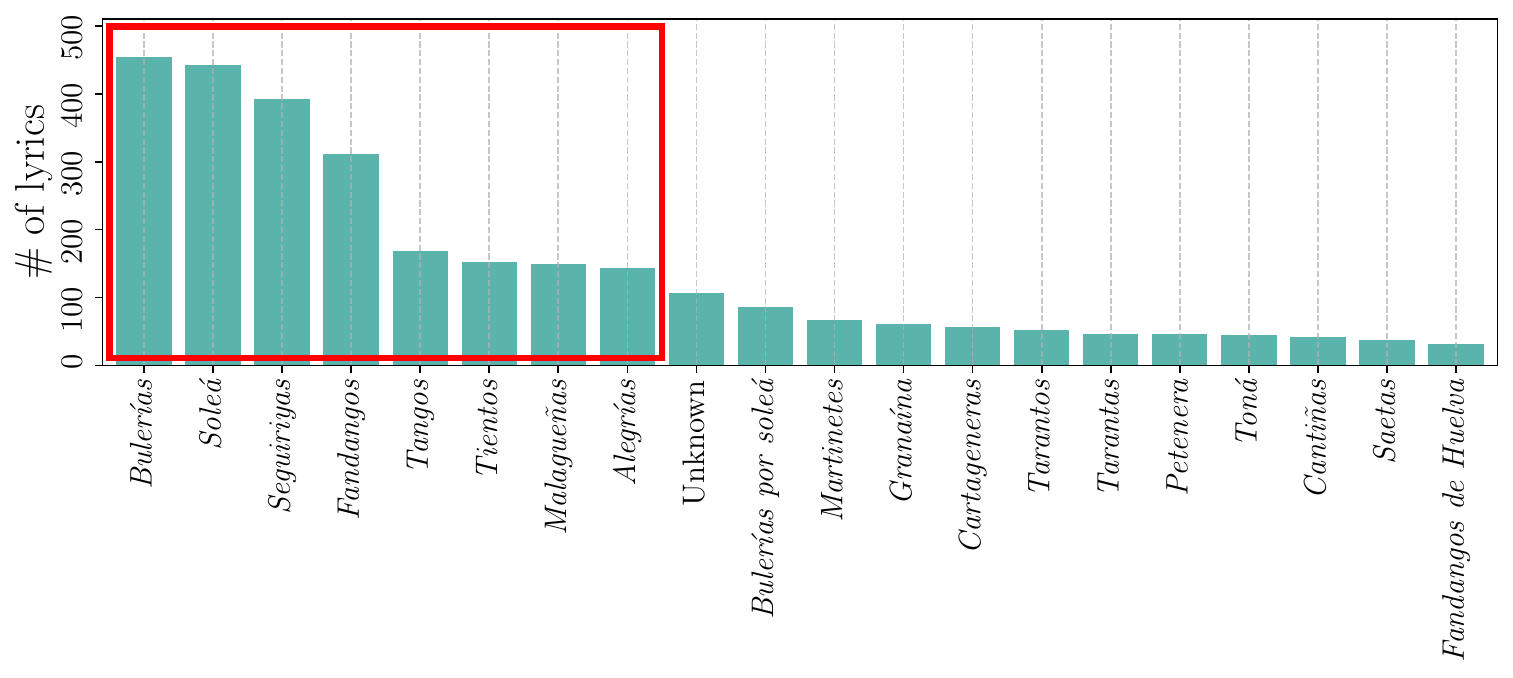}
  \caption{Distribution of the number of lyrics for the 20 most represented genres or \textit{palos} in the corpus. With a red rectangle, we specify the 8 \textit{palos} with the highest representation. There are 58  additional \textit{palos} which count on less than 30 lyrics, which are not shown due to their high under-representation.}
  \label{fig:palohist}
\end{figure}

Also, for the sake of clarity while reading this manuscript, we specify in Table~\ref{tab:palosacronyms} the acronyms with which we will refer to the aforementioned styles throughout the text and in the Figures. It is important to note that categorizing Flamenco songs into styles by ear can be subjective, but for validation purposes, we rely on the classification provided by the album in which each song appears. We consider this classification as the ground truth.

\begin{table}[]
\begin{tabular}{|c|c|}
\hline
\textit{palo}    & Acronym \\ \hline \hline
\textit{alegrías}   & A       \\ \hline
\textit{bulerías}   & B       \\ \hline
\textit{fandangos}  & F       \\ \hline
\textit{malagueñas} & M       \\ \hline
\textit{seguiriyas} & Se      \\ \hline
\textit{soleá}      & So      \\ \hline
\textit{tangos}     & Ta      \\ \hline
\textit{tientos}    & Ti      \\ \hline
\end{tabular}
\medskip
\caption{Acronyms used throughout the text and Figures to refer to the main 8 Flamenco genres or \textit{palos}.}
\label{tab:palosacronyms}
\end{table}

The corpus with the 8 most represented \textit{palos} consists of 2216 song lyrics, 186798 tokens (words), and 10204 types (unique words). 

After preprocessing the corpus, a necessary step previous to their computational analysis which involves deleting punctuation signs and accents and distinguishing between common and proper nouns (see Appendix~\ref{sec:filtering}), we can extract information from the lexical structure of the lyrics. We now proceed to describe several findings in the process of corpus characterization.

\paragraph{Vocabulary size distribution.}

\begin{figure*}[t]
    \centering
    \begin{subfigure}[t]{0.4\linewidth}
        \centering
        \includegraphics[width=\linewidth]{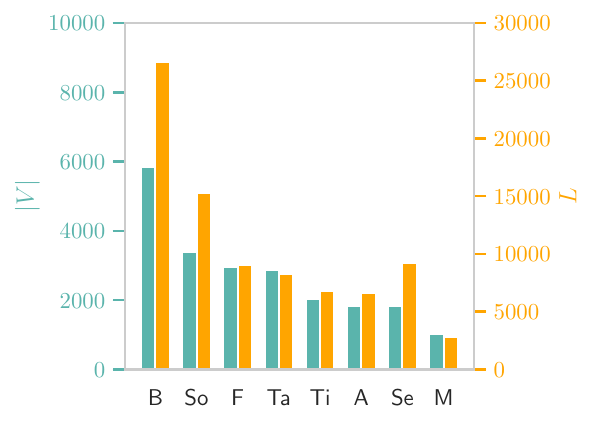}
        \caption{}
        \label{fig:filtered_types_dist_bar}
    \end{subfigure}
    \hfill
    \begin{subfigure}[t]{0.4\linewidth}
        \centering
        \includegraphics[width=\linewidth]{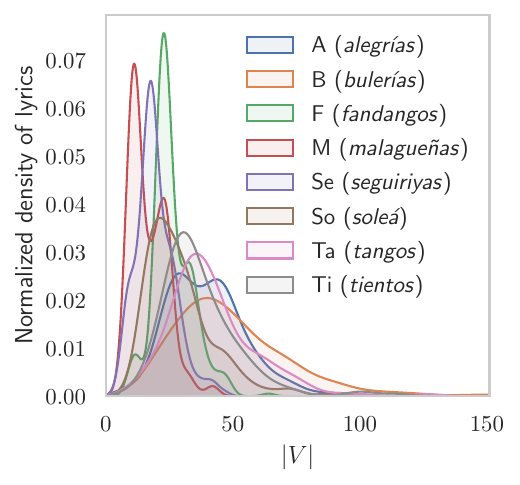}
        \caption{}
        \label{fig:filtered_types_dist}
    \end{subfigure}
    \caption{(\textbf{a}) Amount of tokens $L$ (bars on the right) and types $|V|$ (bars on the left) for each style, and (\textbf{b}) distribution of the vocabulary size, $|V|$ (number of types) for each style. The continuous distribution is derived from a discrete histogram using the kernel density estimation method~\cite{davis2011}.}
    \label{fig:filtered_types}
\end{figure*}

One way to quantify the lexical richness of a text is to measure its vocabulary size $|V|$, i.e., the amount of types. In Fig.~\ref{fig:filtered_types_dist_bar} we show the vocabulary size and the amount of tokens of the eight different styles or \textit{palos}, computed as the vocabulary size of an aggregation of all the lyrics of each \textit{palo}. \textit{Bulerías} account for the highest number of types used while \textit{malagueñas} restrict the range of use to a vocabulary almost 6 times lower in size. The overall ranking of token and type frequencies remains consistent across genres, with the exception of \textit{seguiriyas}.

In the context of song-wise analysis, we may also observe in Fig.~\ref{fig:filtered_types_dist} that \textit{bulerías} have the widest distribution of types used among songs, while \textit{malagueñas}, \textit{fandangos} and \textit{seguiriyas} are restricted to a similar number of types among songs. Cases like \textit{malagueñas} or \textit{alegrías} resemble bimodal distributions, indicating that mainly two different vocabulary sizes characterize their lyrics. With these remarks, we can already infer that \textit{bulerías} serve as the most polyvalent style, with the broadest vocabulary size distribution and the highest amount of types.

\paragraph{Standardized type-to-token ratio.}

Another way of quantifying the lexical richness of a document is to compute its type-to-token ratio, $TTR$, i.e., the ratio between the amounts of types and tokens present in the document~\cite{Malvern2004_1}:
\begin{equation}
\label{eq:ttr}
    TTR = \frac{|V|}{L},
\end{equation}
where $L$ is the number of tokens (or document length). In this way, $1/L \leq TTR \leq 1$. The lower the $TTR$, the richer the document is in terms of the amount of times each type is used on average. 

\begin{figure}[t]
  \centering
  \includegraphics[width=0.75\linewidth, angle=0]{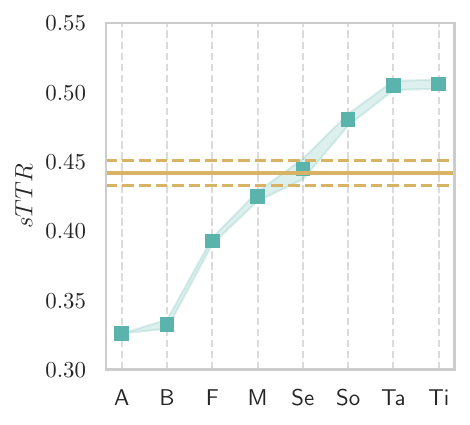}
  \caption{Standardized type-to-token ratio $(sTTR)$ for each genre. With a horizontal continuous line, we show the mean $sTTR$ for the entire corpus (our null model), and with two discontinuous lines, we show the error of the mean.}
  \label{fig:filtered_ttr}
\end{figure}

In our case, Eq.~\ref{eq:ttr} is affected by the fact that some genres use more types than others, as we see in Fig.~\ref{fig:filtered_types_dist_bar}, while also having different amount of tokens and lyrics. For that reason, we use the standardized $TTR$, $sTTR$, which is defined as the $TTR$ averaged over several portions of the same document, with all the portions having the same number of tokens. For each \textit{palo} we create a document compiling all the genre's lyrics. \textit{Malagueñas} is the genre with the least amount of tokens. Then, we take 50 portions of each genre with the length of the entire \textit{malagueñas} subcorpus and compute the mean $sTTR$ and its error except for \textit{malagueñas}, for which we take the entire genre a single time without error. We have checked that increasing the portion number above 50 does not significantly alter our results. We show the results in Fig.~\ref{fig:filtered_ttr}, where the solid line accounts for the $sTTR$ of the entire corpus, our null model for comparison. 

The different values of $sTTR$ provide a partial answer to RQ1 by revealing that various Flamenco genres are distinguished by their unique vocabulary distributions and levels of lexical richness. However, it is not clear that these measurements would allow for an unambiguous identification of a \textit{palo} given some lyrics. For example, the vocabulary distribution is heavily influenced by the specific corpus used, and the reliability of $sTTR$ as a measure of lexical diversity independent of text length has been challenged~\cite{Malvern2004_2}. Moreover, $sTTR$ values for some \textit{palos} are close enough to be indistinguishable, as in the case of \textit{tangos} and \textit{tientos}, or \textit{bulerías} and \textit{alegrías}. To provide a comprehensive answer to RQ1, we need a more sophisticated approach. Thus, we will now examine the specific lexicon utilized by each \textit{palo} using a machine learning algorithm which has proven to be useful in text analysis.

\subsection{\textit{Palo} classification}
\label{sec:results_paloclass}

We employ a Multinomial Naive Bayes (MNB) model \cite{Kibriya2005}, a machine learning classification algorithm, to classify lyrics into distinct Flamenco genres or \textit{palos} solely based on their unigram lexical content. Initially, the corpus is partitioned into a training set comprising 85\% of the data and a validation set containing the remaining 15\%. The algorithm is then trained using the training set, considering the \textit{palo} labels provided in the dataset as the ground truth. Subsequently, the model performance is assessed using the validation set, wherein the genres assigned by the model are compared against the true genre labels associated with the lyrics. Further details about how this classification algorithm works are included in Appendix~\ref{app:mnb}.

\begin{figure}[t]
  \centering
  \includegraphics[width=0.8\linewidth, angle=0]{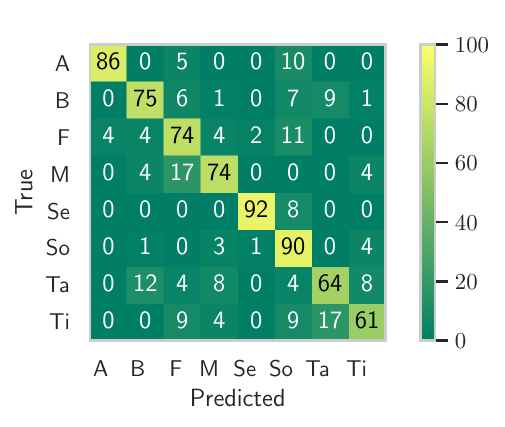}
  \caption{Confusion matrix of the MNB for an arbitrary training, showing the percentage of each \textit{palo} or genre correctly predicted or confused.}
  \label{fig:confmat}
\end{figure}

Let us take a look at a single training of the algorithm. In Fig.~\ref{fig:confmat} we show the confusion matrix of an arbitrary training. In this matrix, we can compare the percentage of lyrics of each genre that have been correctly or incorrectly classified. The true genre is shown in the vertical axis while the classification result is shown in the horizontal axis. Both \textit{seguiriyas} and \textit{soleá} are correctly assigned in 90\% of the times or more, and \textit{alegrías}, \textit{bulerías}, \textit{fandangos}, in at least 75\% of the cases. 

\begin{figure*}[t]
    \centering
    \begin{subfigure}[b]{0.5\linewidth}
      \centering
      \includegraphics[width=1\linewidth, angle=0]{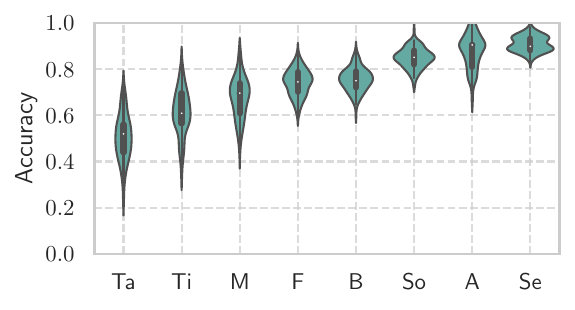}
      \caption{}
      \label{fig:accs}
    \end{subfigure}
    \hfill
    \begin{subfigure}[b]{0.39\linewidth}
      \centering
      \includegraphics[width=1\linewidth, angle=0]{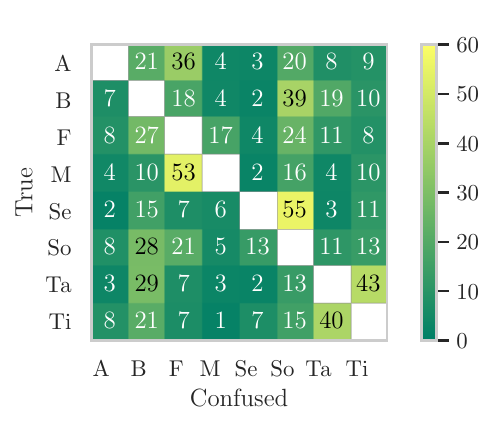}
      \caption{}
      \label{fig:mean_confmat}
    \end{subfigure}
    \caption{(\textbf{a}) Violin plot of the distribution of accuracies obtained for each genre across 100 trainings of the Multinomial Naive Bayes model using various corpus splittings for training and validation, and (\textbf{b}) average confusion matrix of the MNB for 100 different trainings, showing the percentage of confusions of each Flamenco style with other genres. The normalization is based only on the confusions.}
    \label{fig:mean_confmat_accs}
\end{figure*}

Quite generally, different arrangements of lyrics for the training and validation sets yield different results. For this reason, we compute the accuracy of the model (the ratio of correct assignments) for 100 different trainings, obtaining the results shown in Fig.~\ref{fig:accs}.

Despite the fact that genres like \textit{soleá} (85\% mean accuracy), \textit{alegrías} (88\%) and \textit{seguiriyas} (91\%) are classified with outstanding accuracies, a few lyrics are incorrectly classified and therefore confused by the model. To examine these discrepancies, in Fig.~\ref{fig:mean_confmat} we plot the average confusion matrix of 100 different trainings and account only for the incorrect assignments.

Notably, we can identify several interesting confusions of the model. For example, the majority of confusions of \textit{bulerías} (39\%) are with \textit{soleá}. The same confusion may be seen with 40\% of \textit{tientos} being confused with \textit{tangos}, or with 53\% of \textit{malagueñas} being confused with \textit{fandangos}. These particular confusions are enlightening, as according to common wisdom, they are historically related genres. In the case of \textit{soleá} and \textit{bulerías}, the latter is believed to have originated from the acceleration of the former~(\cite{Gamboa2005},~p.~413). Another theory posits a hypothetical common origin for the two genres~\cite{Castro2013}. Similarly, \textit{tientos} are known to be a slow variation of \textit{tangos}, and \textit{malagueñas} are related to \textit{fandangos} according to common wisdom. Remarkably, without any knowledge about the musical features or historical origin of the different genres, the algorithm can discern their relationships considering only their lexical content. We will later delve into these relationships.

It is evident how the aforementioned confusions affect the accuracy of the model, as in the cases of \textit{tangos} (mean accuracy of 50\%) and \textit{tientos} (61\%). However, Fig.~\ref{fig:accs} shows that \textit{soleá}, \textit{alegrías} and \textit{seguiriyas} are classified with outstanding accuracies. Thus, we can state that a classification of Flamenco lyrics into their corresponding \textit{palos} based solely on their lexicon is quantitatively possible (RQ1), and that the inaccuracies are due to interesting \textit{palo} relationships. Now, let us extract the lexical features that allow the model to classify the lyrics, i.e., the characteristic lexicon of each style.

\subsection{Characteristic lexicon extraction}

\begin{figure*}[t] 
    \centering
    \includegraphics[width=\linewidth]{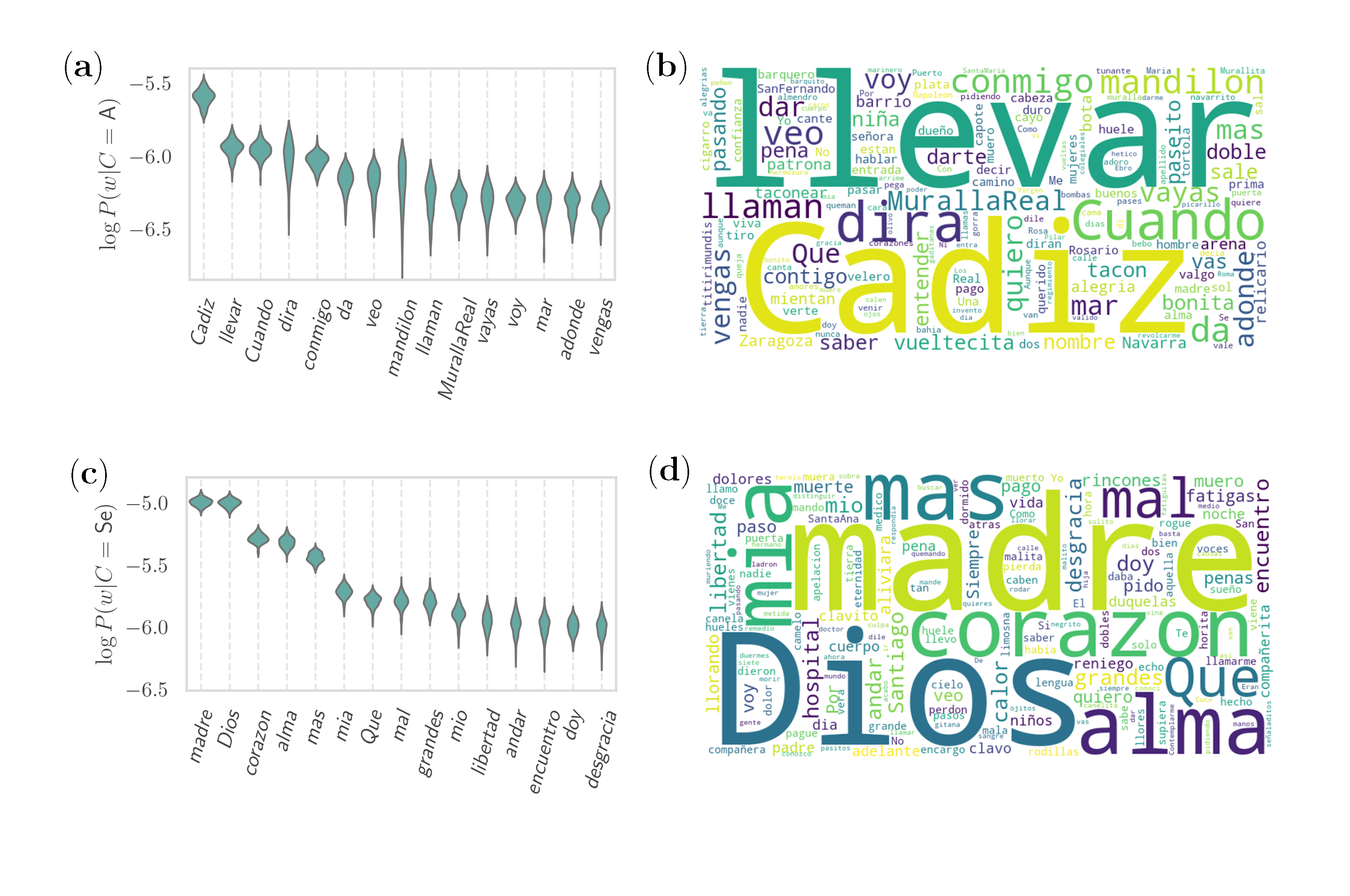} 
    \caption{Summary of the most characteristic words for \textit{alegrías} and \textit{seguiriyas}. We plot the log-probability of the 15 most characteristoc words for (\textbf{a}) \textit{alegrías} and (\textbf{c}) \textit{seguiriyas}. Additionally, we show two wordcloud diagrams in which the words' sizes reflect how characteristic they are for (\textbf{b}) \textit{alegrías} and (\textbf{d}) \textit{seguiriyas}.}
    \label{fig:wide_characteristic_words}
\end{figure*}

The MNB computes a metric called log-probability, $\log P(w|C)$, of each word $w$ for each \textit{palo} $C$ (see Appendix~\ref{app:mnb}). The higher $P(w|C)$ is, the more characteristic $w$ is for $C$. We compute $P(w|C)$ for several trainings of the algorithm and obtain results as shown in Fig.~\ref{fig:wide_characteristic_words}, where we can see the 15 most characteristic words, i.e., the 15 words with the largest value of $\log P(w|C)$, for two example styles, $C = \mathrm{A}$ (\textit{alegrías}) [Fig.~\ref{fig:wide_characteristic_words}(a)] and $C = \mathrm{Se}$ (\textit{seguiriyas}) [Fig.~\ref{fig:wide_characteristic_words}(c)]. 

While \textit{alegrías} feature as most characteristic words references to places such as Cadiz, \textit{Muralla Real} `Royal Wall' or \textit{mar 'sea'} (all related to Cadiz and its harbors), and other less specific words such as \textit{llevar} `to take/bring', \textit{voy} `I go', \textit{conmigo} `with me', \textit{seguiriyas} are biased towards abstract and profound concepts such as \textit{Dios} `God', \textit{alma} `soul', \textit{libertad} `freedom' and \textit{desgracia} `misfortune'. We will later interpret the characteristic words in Sec.~\ref{sec:discussion}, and see how these words clearly differentiate between genres.

\begin{figure}[t]
  \centering
  \includegraphics[width=0.75\linewidth, angle=0]{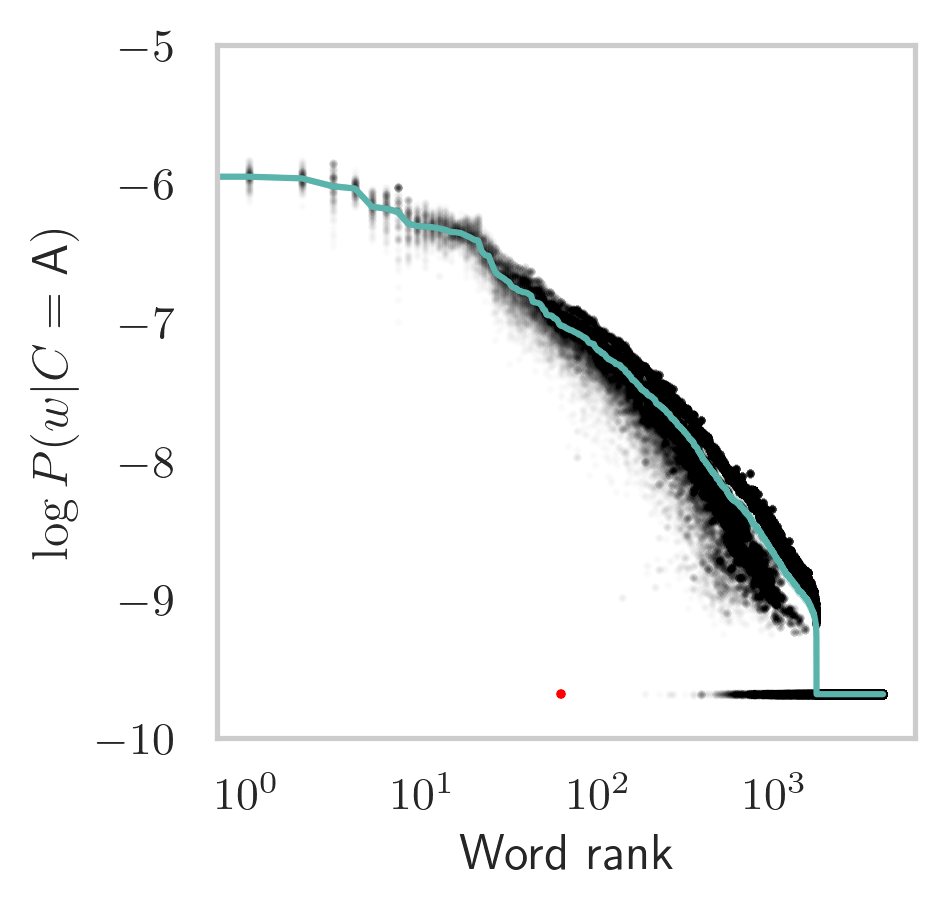}
  \caption{Complete $\log P \left( w|C = A \right)$ distribution over 100 trainings. The words are ranked following the mean value of the $\log P(w|C)$ distribution, which is plotted as a green line. Among the realizations for which words reach the minimum value of $P(w|C)$, the word with the highest rank determines the threshold for the definition of essential words. We indicate the threshold with a red dot. See Appendix~\ref{app:essential} for further details.}
  \label{fig:total_logp_alegrias}
\end{figure}

The vocabularies of different Flamenco \textit{palos} are generally large, consisting of thousands of unique words. In addition, the log-probability distribution for a given Flamenco genre, as seen in Fig.~\ref{fig:total_logp_alegrias} for $C = \mathrm{A}$  (\textit{alegrías}), is smooth and does not allow to retrieve a set of characteristic words. To perform an analysis of the main lexical fields, it is necessary to choose a threshold amount of words, ranked based on their log-probability. However, to make this threshold non-arbitrary, we define a set of essential words for each genre. For a word to be considered essential in a genre, it must fulfill stringent requirements in addition to being among the most characteristic words of the genre. Basically, an essential word maintains its high relevance for a given genre throughout all trainings. Hence, we use the first word with a minimum relevance in at least one training as a threshold. These requirements are explained in detail in Appendix~\ref{app:essential}.

\begin{figure}[t]
  \centering
  \includegraphics[width=0.7\linewidth, angle=0]{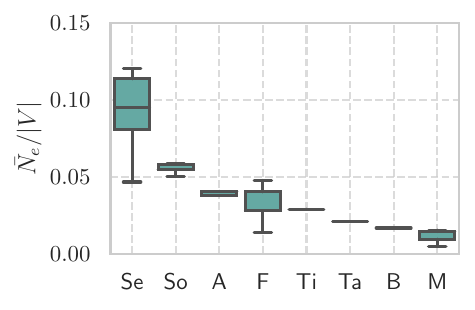}
  \caption{Distribution of the number of essential words for each style or \textit{palo} normalized by the amount of types. Due to the vast number of potential configurations of training sets (approximately $10^{404}$), we sample 500 different distributions of $\log P\left(w|C \right)$ and compute the box plots of essential words. The mean value of each distribution is shown in Table~\ref{tab:esswords}.}
  \label{fig:essential_words}
\end{figure}

The number of essential words within each \textit{palo} serves for distinguishing among various Flamenco genres. Table~\ref{tab:esswords} presents the count of essential words per \textit{palo}, denoted as $N_e$. Notably, while \textit{malagueñas} have 16 essential words, \textit{soleá}'s essential lexicon is significantly larger, averaging 186 words, i.e., nearly 12 times more. Despite this, as illustrated in Fig.~\ref{fig:filtered_types_dist_bar}, \textit{malagueñas} exhibit a far smaller vocabulary compared to \textit{soleá}. To account for the impact of vocabulary size on the essential lexicon, we normalize $N_e$ by the vocabulary size, $|V|$, as depicted in Fig.~\ref{fig:essential_words}. When adjusted for vocabulary size, \textit{seguiriyas} boasts the largest essential lexicon, while \textit{soleá}'s normalized essential lexicon (5.5\%) is only four times greater than that of \textit{malagueñas} (1.5\%). The complete lists of these words for all Flamenco genres in descending order of log-probability are included in Appendix~\ref{app:wordlist}. We emphasize that the relevance of these words is automatically detected and is therefore free of subjective biases. For example, it also identifies certain amount of hapax legomena at \textit{palo} level as significant, as discussed in Appendix~\ref{app:hapaxlegomena}. This is a clear advantage of our application of computational classification methods.

\begin{table}[t]
\begin{tabular}{|c|c|c|}
\hline
\textit{palo}            & $\bar{N_e}$ & $\bar{N_e}/|V|$ \\ \hline \hline
A    & 72 & 0.039 \\ \hline
B    & 98 & 0.017 \\ \hline
F  & 94 & 0.032 \\ \hline
M   & 15 & 0.015 \\ \hline
Se  & 173 & 0.095 \\ \hline
So    & 186 & 0.055 \\ \hline
Ta     & 60 & 0.021 \\ \hline
Ti     & 54 & 0.027 \\ \hline
\end{tabular}
\caption{Absolute and normalized average number of essential words for each Flamenco genre.}
\label{tab:esswords}
\end{table}

Now that we have demonstrated that each genre exhibits its own peculiarity in the lexical distribution and variation of its lyrics, the natural question is RQ2, namely, what relationships exist between the main Flamenco \textit{palos} based on their lexicon? 

\subsection{Relationship between \textit{palos}}

To address RQ2, we quantify the lexical diversity across different \textit{palos}. Hence, we utilize the cosine distance measure of lexical distance between documents (refer to Appendix~\ref{app:distance} for further details and definition), which ranges between 0, indicating a close resemblance between documents, and 1, indicating a significant difference between documents. Our approach involves compiling a single document for each genre by concatenating all its lyrics. Subsequently, we compute the distance between these aggregations.

\begin{figure}[t]
  \centering
  \includegraphics[width=0.8\linewidth]{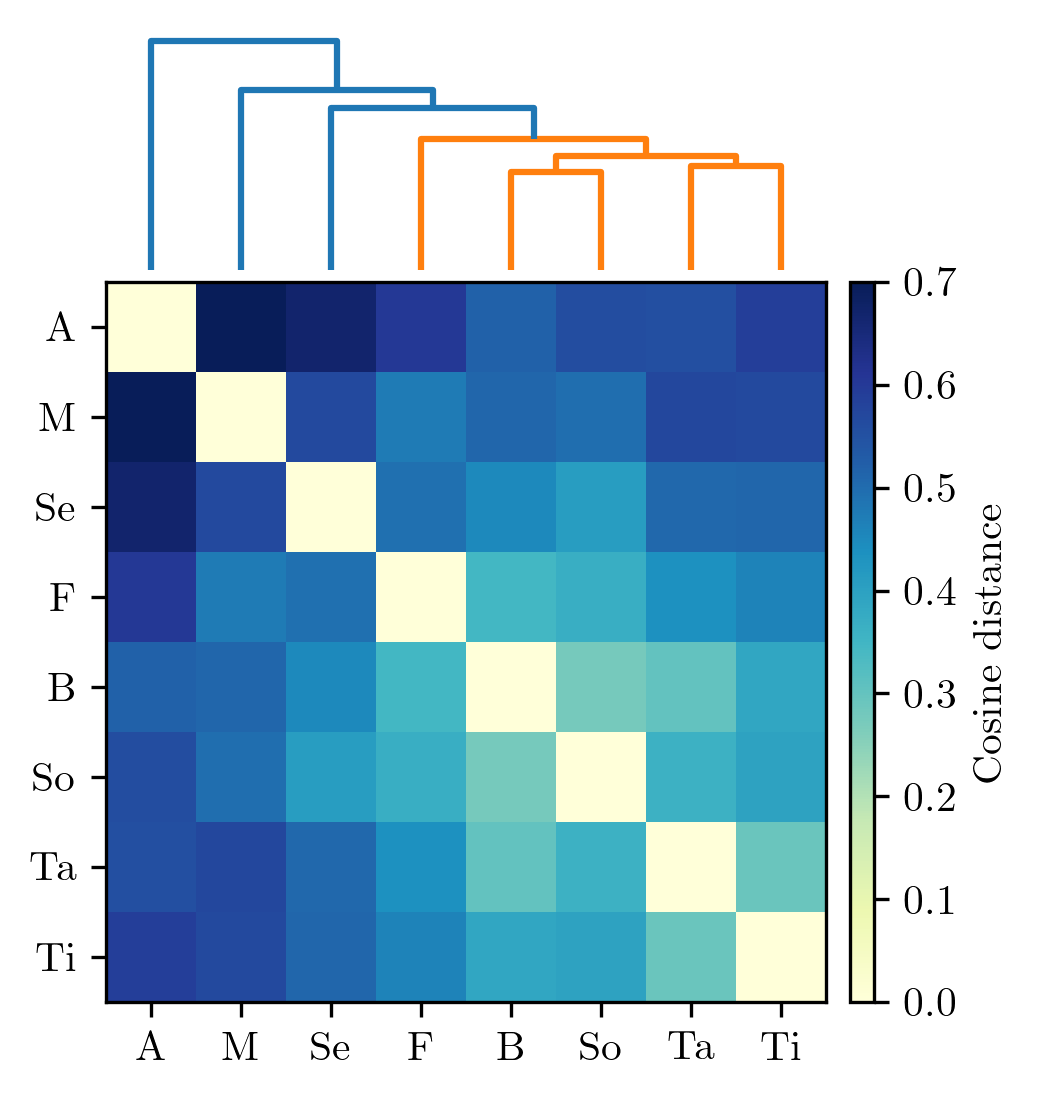}
  \caption{Cosine distance between every pair of different styles or \textit{palos}. The labels on the axis follow the acronyms of Table~\ref{tab:palosacronyms}. Blue color denotes greater distances indicating more distant relationships, while yellow indicates closer relationship. The dendrogram along the top axis illustrates hierarchical clustering distances. Notably, clusters emerge, such as the grouping of \textit{tangos} (Ta) and \textit{tientos} (Ti), as well as \textit{bulerías} (B) and \textit{soleá} (So), due to their common lexical features. It is worth noting that although the cosine distance is theoretically defined within the range of 0 to 1, in our computed distances, the maximum value hovers around 0.7. This suggests that all \textit{palos} exhibit some degree of proximity to one another, albeit to varying extents.}
  \label{fig:palocosdist}
\end{figure}

In Fig.~\ref{fig:palocosdist}, we present the pairwise distance matrix between \textit{palos}, where we find intriguing patterns. For instance, \textit{alegrías} appear to be the most distant style, exhibiting the highest distance from all other styles. In contrast, \textit{tientos} and \textit{tangos} exhibit remarkable proximity with a distance of 0.26, while \textit{bulerías} and \textit{soleá} also demonstrate close associations, with an also small distance of 0.28. The hierarchical clustering revealed by the dendrogram positioned atop the matrix offers insights into these observations. Specifically, it elucidates the amalgamation of \textit{tientos} and \textit{tangos} or \textit{bulerías} and \textit{soleá} into two clusters, which further incorporate \textit{fandangos}, and so forth.

\begin{figure*}[t]
    \centering
    \begin{subfigure}[b]{0.5\linewidth}
      \centering
      \includegraphics[width=1\linewidth]{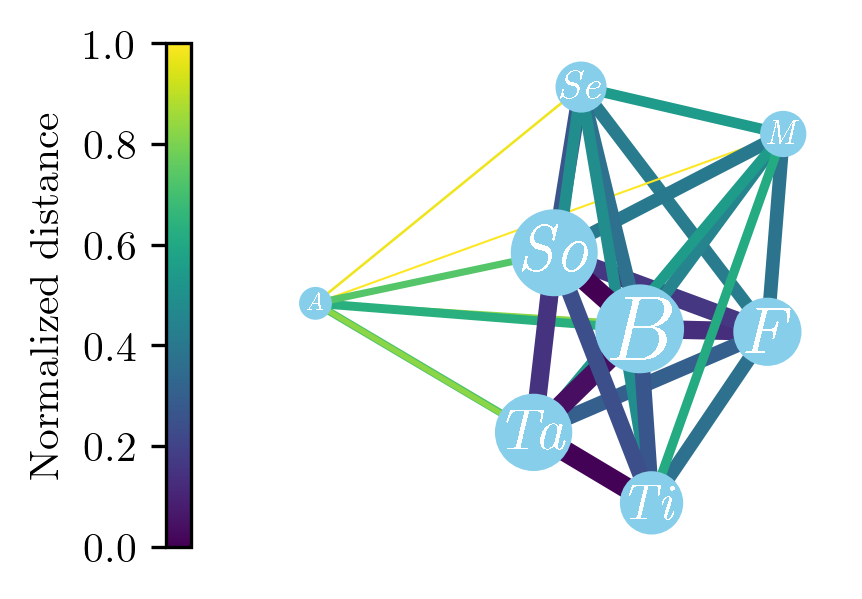}
      \caption{}
      \label{fig:palodisnet}
    \end{subfigure}
    \hfill
    \begin{subfigure}[b]{0.4\linewidth}
      \centering
      \includegraphics[width=1\linewidth]{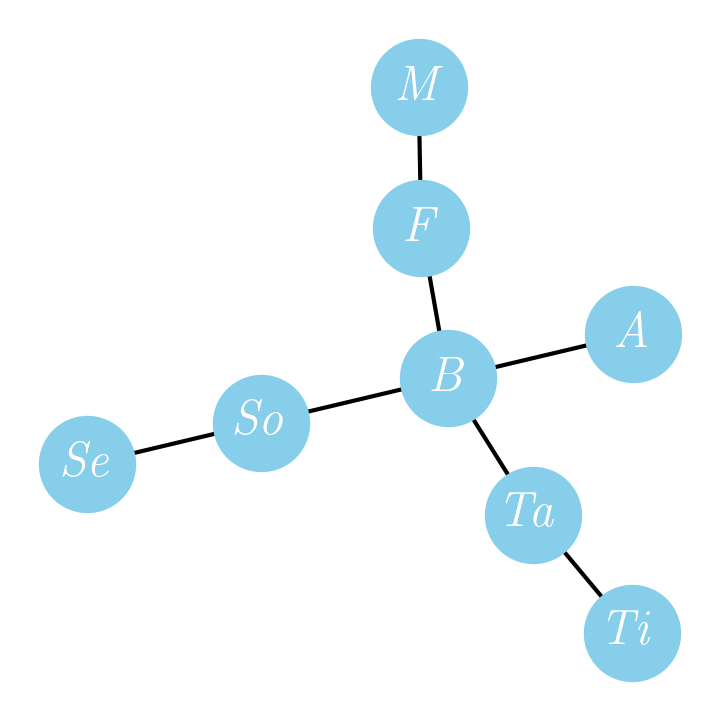}
      \caption{}
      \label{fig:palomst}
    \end{subfigure}
    \caption{(\textbf{a}) Network relationships between different styles or \textit{palos}, with each node representing a \textit{palo} or style. The thickness and color of the edges indicate the distances between the \textit{palos}, and the size of each node is proportional to its closeness centrality. (\textbf{b}) Minimum spanning tree of Flamenco \textit{palos}, showing four branches connected by \textit{bulerías}. The labels on the nodes follow the acronyms of Table~\ref{tab:palosacronyms}.}
    \label{fig:distances}
\end{figure*}

The distances between Flamenco styles are therefore helpful in addressing RQ2. To provide a more concrete analysis, we adopt a network representation. In Fig.~\ref{fig:palodisnet}, we present a fully-connected network illustrating the distances between styles, enabling a straightforward visual assessment of their relationships. Each node represents a style, and the edges vary in thickness and color depending on the cosine distance values. The network structure provides insights into the relationships among different \textit{palos} in several ways. As observed in the dendrogram, \textit{alegrías} emerges as the most lexically distinct \textit{palo}. The remaining \textit{palos} form a more tightly interconnected network, with \textit{bulerías}, \textit{soleá}, and \textit{tangos} prominently contributing to its connectivity. Additionally, the size of the nodes in the network graph corresponds to their closeness centrality, a measure in network analysis that quantifies how close a node is to all other nodes in the network~(\cite{Estrada2015},~p.~146), suggesting their significance in facilitating connections between different musical forms. This network representation proves advantageous as it facilitates the extraction of spanning trees~(\cite{Estrada2015},~p.~27), which are sub-networks that connect all genres such that each genre is linked directly or indirectly to every other genre, without creating any cycles, thereby maintaining the network connectivity. By selecting the minimum spanning tree (MST), which minimizes the sum of distances between \textit{palos}, we obtain the representation depicted in Fig.~\ref{fig:palomst}, i.e., the MST of Flamenco \textit{palos}. This approach enables a structured exploration of the relationships between \textit{palos}, shedding light on their hierarchical organization within the Flamenco genre.

\section{Discussion}
\label{sec:discussion}

By answering RQs 1 and 2 we have proven that Flamenco genres can be distinguished by means of their vocabulary, and also that the study of their lexical variation can establish relationships between different \textit{palos}. We now delve into the meaning of the quantitative results obtained computationally and explore their implications.

At first glance, regarding the $sTTR$ analysis, in Fig.~\ref{fig:filtered_ttr} we can see that $sTTR$ is genre-distinctive. Indeed, returning to the results of the vocabulary size distribution in Fig.~\ref{fig:filtered_types_dist}, if we focus on the $sTTR$ values, we can see how \textit{bulerías} make lower use of their types than \textit{malagueñas} do, even though the vocabulary of the latter is significantly lower than the vocabulary of the former. This indicates that while \textit{bulerías} are rather varied and move within a wide range of topics, this genre makes relatively less use of their topics than \textit{malagueñas} do. Values above the null model's $sTTR$ indicate a more developed use of a specific set of types, while values below indicate either a wider use of types or a less developed use of the available types. 

However, $sTTR$ values do not provide insights into the specific lexical usage across different Flamenco styles. To address this, we employ the MNB model to identify a set of essential words for each genre. As illustrated in Fig.~\ref{fig:essential_words}, the quantity of essential words already varies noticeably across genres. In terms of the number of essential words, \textit{malagueñas} and \textit{bulerías} have the fewest, while \textit{seguiriyas} and \textit{soleá} stand out. Establishing an amount of essential words for each Flamenco style enables us to analyze the topics reflected in each genre's set of essential words.

Flamenco encompasses a variety of genres, each with its own unique origins, tones, and topics. For instance, \textit{alegrías} are typical from Cádiz and are typically fast-paced and celebratory, while \textit{seguiriyas} tend to be slower, more intimate, and melancholic, highly related to the gipsy people's sorrows and sufferings. These distinctions in the essence of various genres are reflected in the variation of their lexicon, as shown in Fig.~\ref{fig:wide_characteristic_words}. 

Before describing the particular features of the essential lexicon of each style, we first identify four primary groups of relevant words (lexical fields) that largely determine the \textit{palo} variation. 

First, we find words related to geographical places and day-to-day life. As previously mentioned, Flamenco lyrics are transmitted orally and encompass a tradition of livings and history. Thus it is common to find references to these topics. Thus, while \textit{alegrías} feature lexicon related to Cádiz, Navarre and \textit{titirimundis} `puppets show', \textit{malagueñas} refer to the \textit{Corte} `Court, Capital' (Madrid), \textit{tangos} talk about wars and \textit{bulerías} feature references to Seville and proper names like Amparo, Dolores or María.

Secondly, we also find references to gipsies and their families and culture, due to the intense influence of this ethnic group in Flamenco art throughout its history. We obtain specific words like \textit{gitano/a} `gipsy', and common words like \textit{madre} `mother', \textit{padre} `father', or \textit{primo/a} `cousin', usually in a gipsy context. Interestingly, \textit{seguiriyas} and \textit{soleá} genres even feature \textit{Caló} vocabulary~\cite{Polackova2011} as characteristic words. The use of \textit{Caló}, a dialect of Romani language spoken by the gipsies of Spain~(\cite{Romani},~p.~125), further connects these genres to the gipsy people.

We find evidence for the belief that Flamenco accompanied gipsies and Andalusian people in times of sorrow and high suffering~\cite{Homann2020} in lexicon related to deep pain, such as \textit{dolor} `pain', \textit{desgracia} `misfortune', \textit{llorar} `to cry', \textit{pena} `sorrow', \textit{traición} `betrayal', \textit{muerta} `dead (singular feminine)'. This semantic field constitutes a third group, along with lexicon related to joy and love. Flamenco is also used to accompany \textit{fiestas} and celebrations, such as the famous gipsy weddings, and thus we may find highly characteristic words such as \textit{adoro} `I adore', \textit{quiero} `I want/love', and \textit{amor} `love'.

Finally, we find a fourth group involving abstract concepts such as \textit{perdón} `forgiveness', \textit{alma} `soul' or \textit{libertad} `freedom' in \textit{palos} like \textit{soleá}, and religious references to God, Christ and to some saints such as \textit{Santiago} `Saint James', \textit{Santa Ana} `Saint Anne', or the Virgin Mary. 

It is relevant to note that the aforementioned lexicon and others participate in formulaic structures or cliches, i.e., fixed expressions that are frequently utilized. Flamenco, being a highly traditional and orally-transmitted art, preserves the structure and lexicon of each style's lyrics, while still allowing for innovation. These fixed expressions constitute coherent stanzas that are typically not semantically connected to the rest of the lyrics but are highly characteristic of a given style and are transmitted from one generation of singers to another. A very recognizable example of a formulaic structure in \textit{alegrías} is 

\begin{center}
\textit{
        a Cádiz no le llaman Cádiz\\~
        que le llaman relicario\\~
        porque tienen por patrona\\~
        a la Virgen del Rosario
}
\end{center}

\begin{center}
`Cadiz is not called Cadiz\\~
but it's called reliquary\\~
because they have as patron saint\\~
the Virgin of the Rosary',
\end{center}
in which the words \textit{relicario} `reliquary' or \textit{patrona} `patron saint' acquire relevance without being semantically connected to the topics of the rest of the lyrics.

Therefore, we face lyrics with a very rich lexical variation. Not only do Flamenco styles feature cultural variation, but also geographical, ethnic, dialectal, and sentimental variation. We now proceed to describe the overall features of the essential words of each Flamenco style. Here, we highlight a selection of essential words from each genre, specifically those having a greater or more intriguing semantic charge. In Appendix~\ref{app:wordlist} we include the complete list of essential words for each \textit{palo}.

Among the essential words of \textit{alegrías} genre, we find terms related to geography such as \textit{Muralla Real} `Royal Wall', \textit{Navarra} `Navarre', \textit{Cádiz} `Cadiz', \textit{San Fernando} `Saint Ferdinand', \textit{mar} `sea', \textit{barrio} `neighborhood', and Zaragoza, the capital of Aragon. This is consistent with the purported origin of \textit{alegrías} stemming from the dynamization of Aragonese \textit{jota} music during the French occupation, with a clear Aragonese influence~(\cite{Gamboa2005} p. 325-236).

\textit{Bulerías} are one of the most gipsy and festive \textit{palos}. With origins in festivity and mockery~\cite{Calado2009}, \textit{bulerías} often feature improvisations at the end of musical performances attended by the gipsy community or in gipsy neighborhoods of Spanish cities. In \textit{bulerías} we find multiple stanzas regarding the virtues of being a gipsy, featuring anecdotes about beauty, kindness, and recurring themes of love, including essential words such as \textit{gitano/a} `gypsy (sing.)', \textit{primo/a} `cousin (sing.)', \textit{quiero} `I want/love', \textit{querer} `to love', \textit{contigo} `with you', \textit{bonita} `beautiful (fem. sing.)'. Another interesting set of essential words is that of female names, such as María, Dolores, or Amparo. These are usually Spanish polysemes -- María may refer to the Virgin Mary, \textit{amparo} `shelter', \textit{dolores} `pains'. The pains, the bad, and the sorrows are typical of \textit{bulerías}, in conjunction with Seville, family, religion, and love.

In \textit{malagueñas} and \textit{fandangos} genres we find references to deep sorrows and religion (\textit{llorar} `to cry', \textit{llorando} `crying', \textit{pena} `pain, sorrow', \textit{traición} `betrayal', \textit{muere} `he/she dies', \textit{morir} `to die'; \textit{muerte} `death' is the most characteristic word of \textit{malagueñas}), but also intensely to the love to women (in \textit{fandangos}, \textit{mujer} `woman' is the most essential word, followed by \textit{quería} `I wanted/I loved' and \textit{querer} `to want/love'), as \textit{fandangos} are usually devoted to sending love messages or love declarations. An interesting lexical feature of \textit{fandangos} is the essential phraseme \textit{cinco sentidos} `five senses', which is directly related to love dependence. As the lover constitutes the five senses of the singer, when the lover is lost, the singer is found suffering and losing the five senses. Also among the essential words of \textit{malagueñas} we find lexical references to desperation and profound emotions, such as \textit{amaba} `I loved' and \textit{delirio} `delirium'. Both \textit{malagueñas} and \textit{fandangos} are referred to as the \textit{cantes} `songs' of Málaga, so it is not surprising that they cover similar lexical fields.

\textit{Seguiriyas} are believed to be, along with \textit{soleá}, the gipsiest and purest Flamenco styles. Indeed, both styles include characteristic references to \textit{Caló} lexicon, such as \textit{duquelas} `fatigues, pains', \textit{camelar} `to desire, consent, enamor', \textit{Undebel} `God', and \textit{terelar} `to have, possess'. \textit{Seguiriyas} and \textit{soleá} also primarily refer to situations of distress and suffering. For this reason, we can find references to sad, deep concepts such as \textit{Dios} `God', \textit{culpa} `guilt', \textit{noche} `night', \textit{ciencia} `science', \textit{infierno} `hell', \textit{muerte} `death', \textit{alma} `soul', \textit{tierra} `earth', \textit{madre} `mother', or \textit{pena} `pain, sorrow'. Interestingly, \textit{seguiriyas} feature specific references to \textit{clavo} `clove', \textit{clavito} `little clove', \textit{canela} `cinnamon', \textit{huele} `smells', and \textit{hueles} `you smell', all of them related to a formulaic expression related to a bath ritual believed in the Romani culture to have beneficial effects in life.

In \textit{tangos}, there are characteristic mentions of geographical locations such as the town of Utrera or the Seville neighborhood of Triana, including its famous \textit{puente} `bridge'. \textit{Dios} `God' is also present, surrounded by other religious references such as \textit{Virgen} `Virgin' or \textit{Humildad} `Humility' in reference to the Christ of Humility. Additionally, around \textit{guerra} `war' we find allusions to historical conflicts, specifically the Cuban and French wars, are viewed through a Spanish historical lens. 

Overall, \textit{tangos} engage deeply with emotions, covering a spectrum from sorrow (\textit{pena} `pain', \textit{fatigas} `fatigues, pains'), to love, happiness and joy (\textit{querer} `to want/love', \textit{bonita} `beautiful (fem. sing.)', \textit{alegría} `joy').

Finally, \textit{tientos} share much of their semantic content with \textit{tangos}, encompassing human emotions and religious themes. This is observed in the use of lexicon such as \textit{querer} `to want/love', \textit{pena} `pain, sorrow', \textit{Dios} `God', \textit{Virgen} `Virgin', and \textit{guerra} `war'. Nonetheless, references to geographical places are less frequent. Predominantly, the essential vocabulary of \textit{tientos} pertains to formulaic expressions, such as \textit{serrana} `mountain woman', \textit{vente} `you come', \textit{conmigo} `with me', \textit{daré} `I will give' in

    \begin{center}
    \textit{
        Vente conmigo serrana\\~
        te daré caña dulce\\~
        que te traigo de La Habana.
       } 
    \end{center}
    
    \begin{center}
    `Come with me, mountain girl\\~
    I'll give you sweet sugar cane\\~
    that I bring from Havana.'
    \end{center}

Vocabulary not only characterizes each Flamenco genre but also establishes relationships among the styles. In Fig.~\ref{fig:palomst} we observe four distinct branches converging at \textit{bulerías}, serving as a unifying node. These branches encompass a gipsy branch involving \textit{seguiriyas} and \textit{soleá}, a branch (presumably) originating from Málaga comprising \textit{malagueñas} and \textit{fandangos}, and the branch of \textit{tangos}, encompassing \textit{tangos} and \textit{tientos}. Importantly, the MST representation effectively encapsulates known historical relationships among \textit{palos}. For instance, our result aligns with established historical accounts such as the derivation of \textit{tientos} from a deceleration of \textit{tangos}~(\cite{Gamboa2005},~p.~338), and sheds light on lesser-known connections, such as the relationship between \textit{malagueñas} and \textit{fandangos}. While the origin of the latter is discussed by various theories~(\cite{Gamboa2005},~p.~420), our finding supports the understanding of current \textit{fandangos} as traditional songs originating from Málaga, thereby linking them to \textit{malagueñas}~\cite{cantesdemalaga}.

In the middle of the four branches, \textit{bulerías} act as a nexus, also allowing \textit{alegrías} to connect to the rest of the \textit{palos}. This may be related to the high vocabulary richness of \textit{bulerías} lyrics and would be interesting to elaborate on in further research. Due to the natural habitat of \textit{bulerías} (entertainment in gipsy meetings, festive ending of Flamenco \textit{fiestas}) this genre could be drawn towards using a wider and richer vocabulary, leading to act as a nexus to the rest of the \textit{palos}.

\section{Conclusions}
\label{sec:conclusions}

To sum up, we have quantitatively analyzed the lexical variations in Flamenco lyrics using computational methods for text analysis and machine learning for classification. Our work constitutes a proof of concept to quantify lexical features of Flamenco lyrics and subsequently relate them to history and culture.

First, we have addressed RQ1, \textit{Can Flamenco songs be properly assigned to their corresponding} palo \textit{solely by means of their lexicon? If so, what is the characteristic lexicon of each} palo\textit{?} Our findings demonstrate that, with varying degrees of accuracy, it is indeed possible to automatically classify Flamenco songs into their respective \textit{palos} based solely on their lexicon. This discovery holds significant relevance for Flamenco \textit{palo} classification, which has traditionally relied on subjective assignments by authors or record labels, guided by stylistic conventions. Our findings quantitatively demonstrate that these conventions are also reflected in consistent lexical patterns.

The fact that \textit{palo} classification is possible based only on the content of their lyrics shows that lyrics of different \textit{palos} encode different information, due to their different origins, history, and habitats. By performing a discussion and interpretation of the \textit{essential} lexicon of each \textit{palo}, we have identified insightful differences between Flamenco styles. While \textit{alegrías} are mainly festive and referent to Cádiz, \textit{seguiriyas} cover deep concerns and embody gipsy terms, while \textit{bulerías} covers a wide range of topics and vocabulary maybe due to its use as an instrument of anmmenization and celebration. 

In virtue of this differentiative characteristic lexicon, we have answered RQ2, \textit{what relationships exist between the main Flamenco} palos \textit{based on their lexicon?}, computing a lexical distance between \textit{palos}. With this metric we have been able to identify four different branches of Flamenco constituted mainly by gipsy styles, \textit{cantes} (singings) from Málaga, \textit{tangos}, and \textit{alegrías}, with a common nexus in \textit{bulerías}. Our Minimum Spanning Tree representation allows us to quantitatively support historical theories on the relationship of \textit{malagueñas} and \textit{fandangos} or between \textit{tientos} and \textit{tangos}.

Finally, our study is not without its limitations.  Unlike mainstream genres like reggaeton, pop, or rock, Flamenco's limited availability of large lyric databases complicates analysis, further compounded by its rich collection of styles. While our study benefited from a reasonably curated corpus, the dataset still lacked representation of numerous \textit{palos}, meaning our conclusions are only reliable for the eight main \textit{palos} analyzed. Furthermore, within the genres that were sufficiently represented, there were significant disparities in representation, with some \textit{palos} more heavily represented than others. This highlights the corpus-dependence of our analysis and conclusions. Future research should address these limitations by expanding data collection efforts, which would strengthen the robustness and generalizability of the findings. With a more comprehensive dataset, more advanced techniques such as BERT, Word2Vec, or Convolutional and Recurrent Neural Networks, could be employed. For example, training an sBERT model on Flamenco lyrics could capture word relationships at a deeper level than our machine learning approach. Currently, Spanish language models do not fully
capture the regional variants often found in Flamenco lyrics, which often differ from the standard Spanish found in typical training datasets. We are hopeful that \texttt{PlanTL-GOB-ES/roberta-base-bne}~\cite{robertabne} and  \texttt{BSC-LT/mRoBERTa}~\cite{mroberta} models will address this gap. Future work should focus on applying these advanced technologies with properly trained datasets.

While our unigram approach has already uncovered complex formulaic structures, further work could explore $n$-grams with $n \geq 2$. This would capture both semantic and syntactic relationships that extend beyond the individual meanings of unigrams, and would allow for the automation of the concatenation process shown in Sec.~\ref{sec:filtering}.

Moreover, a change in the preprocessing stage, stop words, or other parameters of our analysis may yield different results, but we believe that the main conclusions of the current study will remain valid. Also, incorporating advanced text processing techniques such as lemmatization or stemming could refine our analysis~\cite{Balakrishnan2014} and yield deeper insights into the lexical nuances of Flamenco lyrics. As Flamenco continues to evolve, marked by contemporary references like Moneo's mention of the coronavirus in recent performances~\cite{moneovideo}, ongoing research efforts must adapt to capture and interpret these dynamic shifts in the genre's lexical content, therefore allowing for a diachronic analysis of the lexicon used in Flamenco lyrics. To sum up, embracing quantitative methods and interdisciplinary applications have allowed us to uncover Flamenco's linguistic and cultural legacy, thereby deepening our appreciation for this iconic musical tradition.

\begin{acknowledgements}
The authors thank Norman P. Kliman, Jos{\'e} Mar{\'\i}a Vel{\'a}zquez-Gaztelu and Joaqu{\'\i}n Guindo for useful discussions and references.
This work was partially supported by the Spanish State Research Agency
(MICIU/AEI/10.13039/501100011033) and FEDER (UE) under project APASOS (PID2021-122256NB-C21) and the Mar{\'\i}a de Maeztu project CEX2021-001164-M, and by the Government of the Balearic Islands CAIB fund ITS2017-006 under project CAFECONMIEL (PDR2020/51).
\end{acknowledgements}

\bibliographystyle{ACM-Reference-Format}
\bibliography{biblio}

\appendix 

\section{Corpus characterization methods}
\label{app:corpus}
\input{app_corpus_resubmission_1}

\section{Additional methods}
\label{app:methods}
\input{app_methods_resubmission_1}

\section{Essential words and lexical fields}
\label{app:wordlist}
\input{app_esswords}

\section{List of stop words}
\label{app:stopwords}
\input{app_stopwords}

\end{document}

%% file: app_corpus_resubmission_1.tex
Previous to the analysis of the lexicon used in the different Flamenco genres, we characterize the corpus~\cite{corpus}.

\subsection{Empirical laws}
\label{sec:emphapax}

\begin{figure*}[t]
    \centering
    \begin{subfigure}[b]{0.45\linewidth}
      \centering
      \includegraphics[width=0.8\linewidth, angle=0]{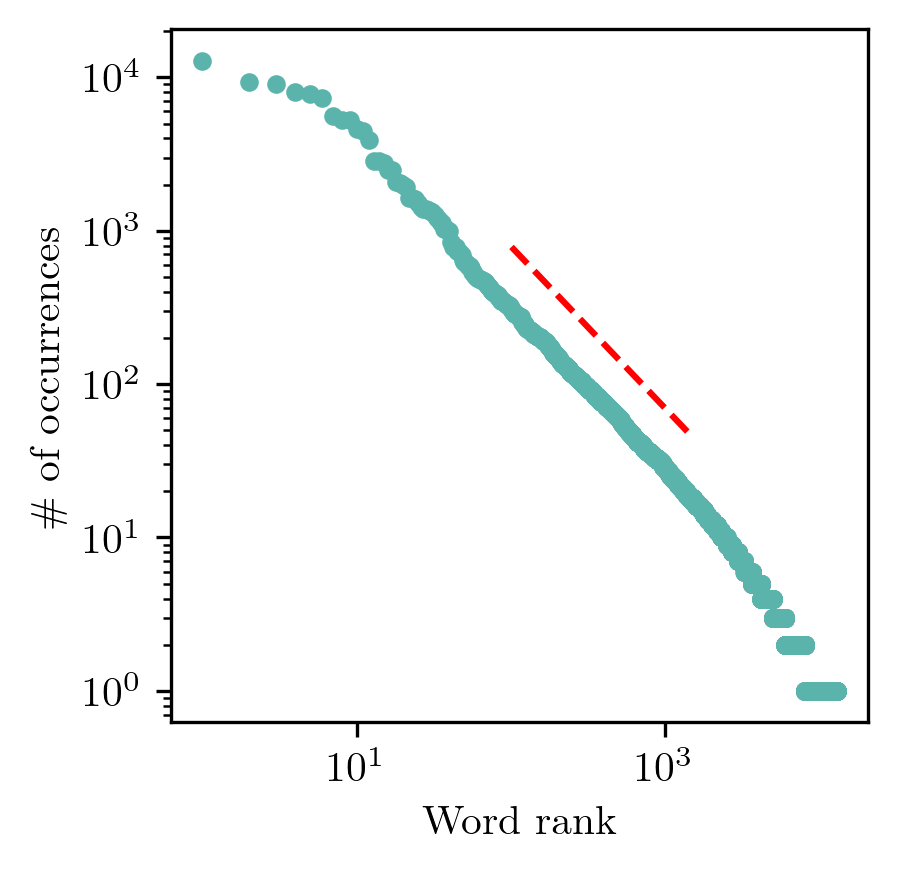}
      \caption{}
      \label{fig:zipf}
    \end{subfigure}
    \hfill
    \begin{subfigure}[b]{0.45\linewidth}
      \centering
      \includegraphics[width=0.8\linewidth, angle=0]{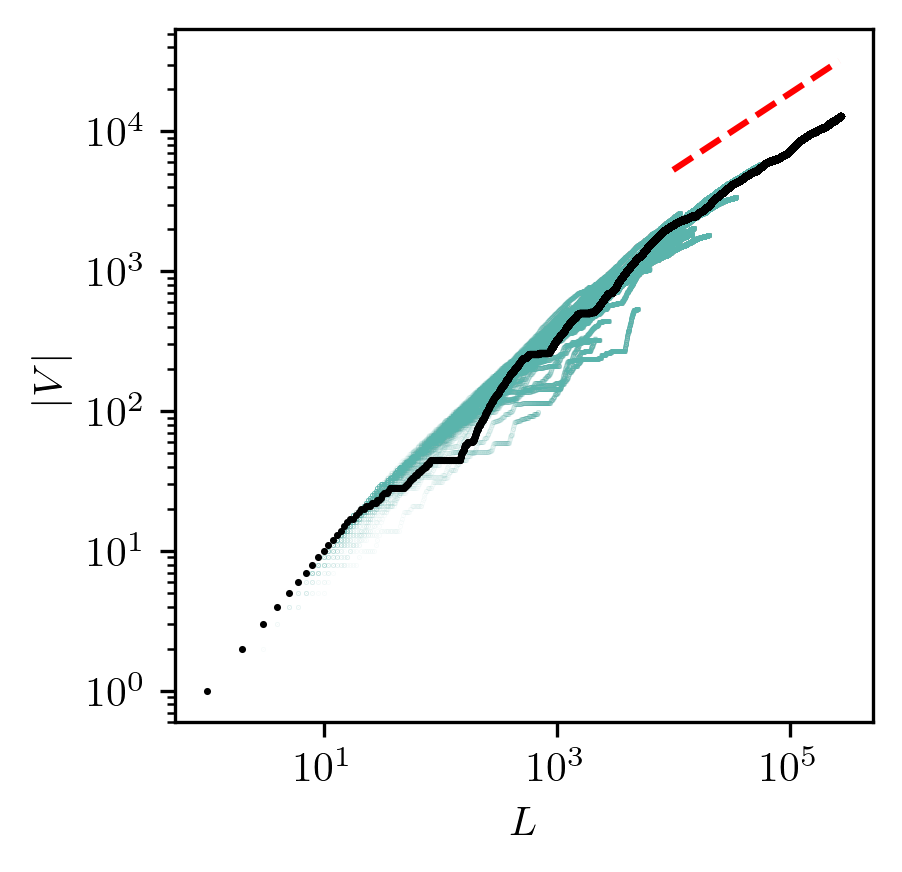}
      \caption{}
      \label{fig:heaps}
    \end{subfigure}
    \caption{(\textbf{a}) Log-log scale distribution of tokens in the entire and unfiltered corpus, fitted to Zipf's law, which relates the number of occurrences of a word with its rank as a power law with an exponent of~$-1.06$ (dashed line), and (\textbf{b}) log-log scale of the relationship between corpus size, $L$, and vocabulary size, $|V|$, adhering to Heaps' law. We plot the relationship for each genre independently (green lines) and for the global corpus (thick black line), in both cases for unfiltered texts and for a given disordered realization. The sublinear scaling observed underscores the increasing lexical diversity as the corpus grows, with an exponent of $0.55$ for large amounts of tokens (dashed line).}
    \label{fig:st_laws}
\end{figure*}

We check its adherence to both Zipf's and Heaps' laws, as illustrated in Fig.~\ref{fig:st_laws}. The observed exponents agree with many other corpora widely employed in the literature~\cite{Altmann2016, Stanisz2024}.

\subsection{Text preprocessing}
\label{sec:filtering}

Before the data analysis, we filter the text of the documents to get rid of elements that do not contribute to our lexical analysis, such as punctuation signs or articles. In this way, we perform the following filtering steps.

\begin{itemize}

    \item We concatenate words that change their meaning when separated, particularly geographical or religious names. For instance, "Santa Ana" becomes "SantaAna," and the city "Jerez de la Frontera" becomes "JerezdelaFrontera," among other instances.

    \item We lowercase certain words that appear in uppercase. This decision stems from the observation that in song lyrics, not only proper nouns are capitalized, but also words that mark the beginning of a verse. To differentiate between proper nouns and verse beginnings, we employ the following criterion. Let $w$ be a word that appears both in uppercase and lowercase in song lyrics. Denote by $n_w$ the number of times the word $w$ appears in lowercase and $N_w$ the number of times it appears in uppercase. If $N_w < \gamma \left( n_w + N_w \right)$, where $\gamma$ is a fixed threshold (in our case, $\gamma = 0.2$), we convert all occurrences of $w$ to lowercase. Otherwise, no changes are made. However, the lowercasing approach has its drawbacks and might fail to lowercase certain words that should be in lowercase. 
    
    \item Because some lyrics were incorrectly accentuated due to manual transcription, we remove the accentuation from all words for the analysis. However, we restore the proper accentuation when interpreting and presenting the results. In addition, we remove all the punctuation forms, i.e., 
    \begin{verbatim}
  ,;.:¡!¿?@#\$%&[](){}<>~=+\-*/|\\^`"'.
    \end{verbatim}

    \item We eliminate stop words from the analysis. Stop words are words that lack inherent meaning in our context and typically include articles, pronouns, adverbs, auxiliary verbs, and exclamations such as \textit{olé}. A comprehensive list of the stop words removed during pre-processing is provided in Appendix~\ref{app:stopwords}. Note that our automatic lowercasing method causes that words such as \textit{Que} 'that', \textit{Una} 'One (fem.)' and \textit{Ay} might not be appropriately removed even though they are part of the stop words list in their lowercased form.
   
\end{itemize}

%% file: app_methods_resubmission_1.tex
During our analysis of Flamenco lyrics, we have employed various computational tools. This section elaborates on the methodologies utilized, encompassing machine learning model implementation, definition and identification of essential words, and computation of lexical distances between documents.

\subsection{Multinomial Naive Bayes} 
\label{app:mnb}

Our goal is to extract the characteristic lexicon of each of the 8 Flamenco \textit{palos} that we have in our corpus. For that, we train a machine learning model called Multinomial Naive Bayes (MNB) classifier with Lidstone smoothing~\cite{Kibriya2005, MNBScikit}. In the following section we include an explanation of the MNB algorithm, only for completeness, as this is a standard approach.

\paragraph{MNB} First, to feed the MNB, we represent the documents with a term-frequency inverse-document-frequency (TF-IDF) matrix instead of a word-count matrix. We do this because the TF-IDF metric accounts not only for the frequency of use of a given word in the corpus but also for how scarcely (or predominantly) the word is distributed among documents. Additionally, using TF-IDF values with the MNB is known to increase its accuracy~\cite{Kibriya2005}. The TF-IDF value for the word type $w$ in a document $d$ of a corpus $D$ is given by

\begin{equation}
    \text{TF-IDF}(w, d) = \bar{tf}(w,d) \left[ 1+ \log\left( \frac{|D|}{df(w)} \right)\right],
\end{equation}
where

\begin{itemize}
    \item $\bar{tf}(w,d)$ is the normalized frequency of word $w$ in document $d$,
    \item $|D|$ is the cardinality of the set of documents, i.e., the amount of documents that constitute the corpus,
    \item $df(w)$ is the document frequency of word $w$, i.e., the amount of documents in which $w$ appears.
\end{itemize}

With the TF-IDF values, we create a matrix with dimensions $(|D|, |V|)$, where $V$ is the vocabulary of the corpus, i.e., the set of unique words present throughout the corpus. In addition, we normalize the TF-IDF matrix so that the Euclidean norm of each row (each document) is~1. This representation yields, then, $|D|$ normalized vectors in an $|V|$-dimensional space. Each vector represents a document, and these vectors are what we feed the MNB model.

To perform the classification, the MNB model first assigns a probability of a term $w$ to pertain to a class $C$, namely $P(w | C)$. In our case, terms are types and classes are Flamenco styles or \textit{palos}. In this way, for any $w$ in our corpus, the MNB model computes a $P(w | C)$, for every \textit{palo} $C$, given by

\begin{equation}
\label{eq:pwc}
    P(w|C) = \frac{\alpha + \sum\limits_{d \in C} \text{TF-IDF}(w,d)}{\sum\limits_{w' \in V}\left( \alpha + \sum\limits_{d \in C} \text{TF-IDF}(w',d) \right)},
\end{equation}
where $\alpha$ is a fixed smoothing parameter which will be discussed later.

With this information, and given an unknown document $d$, the MNB model assigns a style $C$ to $d$ based on a score computed as
\begin{equation}
\label{eq:prob}
    \text{Score}(C|d) = \log[P(C)] + \sum_{w \in d} \left(\log \left[P(w | C)\right] \text{TF-IDF}(w,d)\right),
\end{equation}
where $P(C)$ is the prior probability of style $C$ estimated from the corpus as the normalized frequency of lyrics of style $C$. In Table~\ref{tab:pcs} we show the different values of $P(C)$ for different \textit{palos}.

\begin{table}[]
\begin{tabular}{|c|c|}
\hline
\textit{palo} & $P(C)$ \\ \hline \hline
A (\textit{alegrías}) & 0.06 \\ \hline
B (\textit{bulerías}) & 0.20 \\ \hline
F (\textit{fandangos}) & 0.14 \\ \hline
M (\textit{malagueñas}) & 0.07 \\ \hline
Se (\textit{seguiriyas}) & 0.18 \\ \hline
So (\textit{soleá}) & 0.20 \\ \hline
Ta (\textit{tangos}) & 0.08 \\ \hline
Ti (\textit{tientos}) & 0.07 \\ \hline
\end{tabular}
\caption{Prior probabilities for the different Flamenco styles, computed as the normalized frequency of documents of each \textit{palo} in the corpus.}
\label{tab:pcs}
\end{table}

Finally, the MNB assigns to the document $d$ the \textit{palo} $C$ with the highest score. 

We now proceed to describe how to assess the actual performance of the model when classifying new documents.

\paragraph{Assessment of the classification performance.} 
We divide our corpus in two sets of documents, the training and the validation sets. The training set is used by the model to learn the characteristic lexicon of each class, i.e., to compute Eq.~\ref{eq:pwc}. This is the training process. Once the model is trained, its performance is assessed by computing Eq.~\ref{eq:prob} for each document of the validation set and comparing these scores with the actual class to which the documents pertain. Let us consider two styles P1 and P2. If the model classifies as P1 certain lyrics pertaining to P1 style, this is a correct assignment. On the contrary, should the actual style of the lyrics be P2, this will be a confusion of the model. With this information we build a confusion matrix as the one in Fig~\ref{fig:confmat}.

The metric used to perform the assessment is called accuracy, and refers to the ratio of correct assignments, i.e., the number of correct classifications divided by the total amount of classifications.

There are several degrees of freedom that affect the accuracy of the model. One of them is the size and composition of the training and validation sets. We configure the size of the training and validation sets to be 85\% and 15\% of the whole corpus, respectively, forcing the distribution of styles in each set to be equal and representative of that of the whole corpus. The model then learns with 85\% of the lyrics, and we assess its performance on the remaining 15\%. The impact of the training and validation set compositions on accuracy is obscured by averaging relevant measures such as the accuracy itself across multiple training sessions with varying training and validation set compositions.

Another parameter that might affect the accuracy is $\alpha$ (see Eq.~\ref{eq:pwc}). We now proceed to show how we optimize the training process using $\alpha$.

\paragraph{Training optimization.} $\alpha$ avoids the zero-frequency problem~\cite{Kibriya2005}. To look for the optimal value of $\alpha$, we perform several trainings and validations of the model for different values of $\alpha$ given a chosen training size of 85\% of the corpus which respects the global distribution of lyrics within styles (see Fig.~\ref{fig:palohist}). We look for maximum accuracy.

\begin{figure}
  \centering
  \includegraphics[width=0.7\linewidth, angle=0]{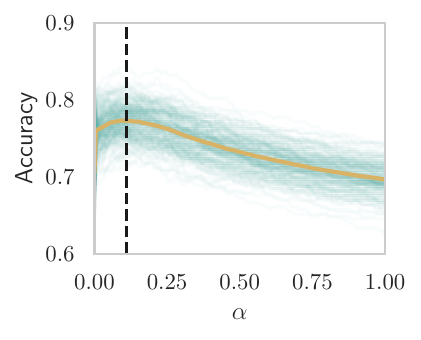}
  \caption{Global accuracy achieved by the model in terms of the value of the smoothing parameter $\alpha$. In green, we show individually 200 different trainings, while in thick orange we show the average for each value of $\alpha$. This average has a maximum of 0.77 in $\alpha = 0.11$.}
  \label{fig:acc_alpha}
\end{figure}

For 200 different trainings, we iterate $\alpha \in [0, 1]$, in steps of $0.005$. On average, the maximum global accuracy, i.e., the maximum ratio of correct assignments in the entire training set, is achieved for $\alpha = 0.11$, as shown in Fig.~\ref{fig:acc_alpha}.

With these values of $\alpha$ and training size, the model behaves as shown in the confusion matrix of Fig.~\ref{fig:confmat} and in the accuracy ranking of Fig.~\ref{fig:accs}. The accuracies are reasonably good for most of the styles. A few styles do not perform as well, due to lexical similarities which are discussed in Section~\ref{sec:results}.

\subsection{Essential words}
\label{app:essential}

In virtue of $P(w|C)$, defined in Eq.~\ref{eq:prob}, each style has a ranking of words, as there are words which have a larger conditional probability of pertaining to a given style than others.

For each random set of training lyrics we will obtain a different set of $P(w|C)$ for a given style $C$. If we compute the mean for a set of trainings we obtain a decreasing curve as a function of the rank of $w$. In Fig.~\ref{fig:total_logp_alegrias} we show as an example the $P(w|C)$ distribution for \textit{alegrías}.

As we rank words for each style based on the mean value of the different $P(w|C)$ computed over multiple training iterations, we observe that, for each iteration, the minimum probability is obtained for several words. This occurrence arises because these words fail to appear in a sufficient number of lyrics to be consistently present across all random sets of training data. To address this issue and identify a set of characteristic words for each style, we rank words based on their average $P(w|C)$ and look for the first word that had the minimum $P(w|C)$ in at least one realization (see red dot in Fig.~\ref{fig:total_logp_alegrias}). The rank of that word defines a threshold, and we term all words ranked above the threshold as essential.The quantity of essential words per style is depicted in Figure~\ref{fig:essential_words}. There, it is clear that different styles have significantly different amounts of essential words. The complete list of essential words for each style is provided in Appendix~\ref{app:wordlist}.

\subsection{Hapax legomena}
\label{app:hapaxlegomena}

Additionally, let us examine the distribution of the ratio of \textit{palo}-hapax legomena, i.e., the set of words that appear only in one \textit{palo} and not in the others, $r_\mathrm{h}$, for each song of each \textit{palo},
\begin{equation}
    r_\mathrm{h} = \frac{\text{\# \textit{palo}-hapax}}{|V_\mathrm{s}|},
\end{equation}
where $|V_\mathrm{s}|$ is the size of the vocabulary of a song.

\begin{figure}[t]
  \centering
  \includegraphics[width=\linewidth]{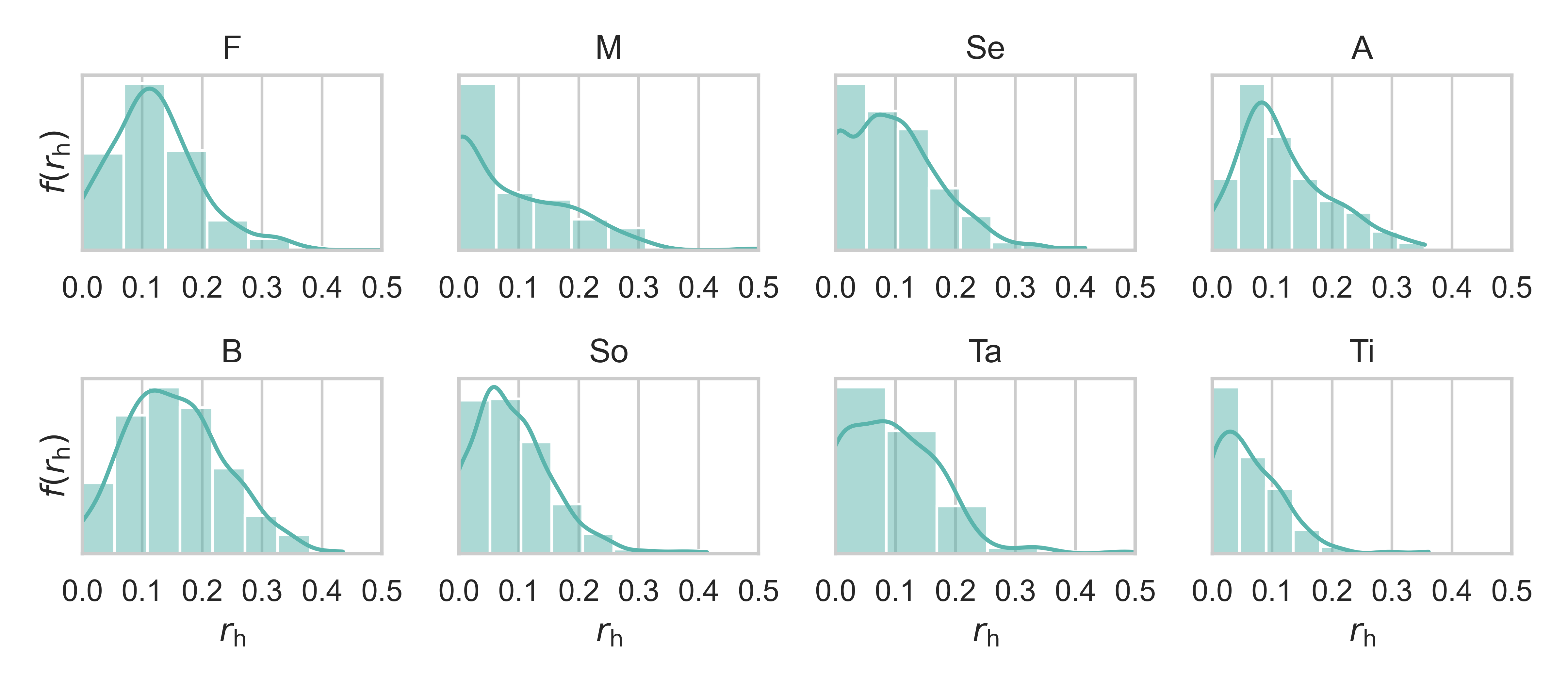}
  \caption{Kernel density estimation~\cite{davis2011} of the distribution of hapax legomena found in the filtered lyrics of each \textit{palo}, following the acronyms of Table~\ref{tab:palosacronyms}.}
  \label{fig:hapax_legomena}
\end{figure}

In Fig.~\ref{fig:hapax_legomena} we show the distribution of $r_\mathrm{h}$ for the filtered dataset. While distinct patterns may be observed, the distributions across all \textit{palos} are generally quite similar. Overall, all distributions have their peaks located around $r_\mathrm{h} = 0.1$, which is lower than $r_\mathrm{h} = 0.5$, the pattern commonly observed in large corpora for corpus-level hapax legomena~\cite{baayen1996, rosillo2024}.

Hapax legomena at \textit{palo} level are considered as relevant in certain ocasions by the MNB classifier fed with TF-IDF scores. Indeed, if we compare the \textit{palo}-unique lexicon with the essential words computed in our manuscript, we find several coincidences, as seen in Table~\ref{tab:genre_ess_shared}.

\begin{table}
\begin{tabular}{|c|c|c|c|}
\hline
\textit{palo} & Genre-unique & Essential & Shared \\ \hline \hline
A   & 376 & 73 & 5 \\ \hline
B   & 2288 & 98 & 0 \\ \hline
F   & 747 & 94 & 1 \\ \hline
M   & 135 & 15 & 1 \\ \hline
Se  & 379 & 172 & 5 \\ \hline
So  & 746 & 182 & 0 \\ \hline
Ta  & 624 & 57 & 0 \\ \hline
Ti  & 273 & 55 & 0 \\ \hline
\end{tabular}
\medskip
\caption{Amount of genre-unique lexicon and essential words for each \textit{palo}, including shared word totals.}
\label{tab:genre_ess_shared}
\end{table}

\subsection{Distance and clustering}
\label{app:distance}

To compute the similarity between different styles we use cosine distance. This is a measure used to quantify the similarity between two document vectors in a multidimensional space. In our case, our vectors are the normalized rows of the TF-IDF matrix, as detailed in Section~\ref{app:methods}.

If we consider two document vectors $\mathbf{T_1}$ and $\mathbf{T_2}$, the cosine distance (CD) between these vectors is calculated using the cosine of the angle between them in their multidimensional space, as
\begin{equation}
    \text{CD}(\mathbf{T_1}, \mathbf{T_2}) = 1 - \mathbf{T_1} \cdot \mathbf{T_2},
\end{equation}
where $\mathbf{T_1} \cdot \mathbf{T_2}$ is the dot product between vectors $\mathbf{T_1}$ and $\mathbf{T_2}$.

CD ranges between 0 and 1. A value of 0 indicates that the two vectors are identical (perfect similarity), while a value of 1 implies that the vectors are orthogonal (completely dissimilar).

In the context of analyzing Flamenco lyrics, CD can be used to compare the lexical similarity between different styles. By representing each style compilation, i.e., an aggregation of all the lyrics of a given style, as a document vector of TF-IDF values, we can calculate the cosine distance between pairs of styles to determine their similarity.

Once a distance metric between styles has been established, we proceed with hierarchical clustering \cite{Murtagh2012} to organize them into clusters. Given the presence of 8 styles, we initialize the process with the 8 aggregations. The hierarchical clustering algorithm begins by assigning the first cluster to the two closest documents. Subsequently, this formed cluster is treated as a single document, and the distance matrix is updated accordingly. This iterative process continues until only one cluster remains. The dendrogram positioned atop the distance matrix in Fig.~\ref{fig:palocosdist} visually depicts this hierarchical clustering process.

%% file: app_esswords.tex
In this Appendix we show the complete list of essential words for each palo, including their $\log P(w|C)$ distribution.

\subsection{\textit{Alegrías}}

In Fig. \ref{fig:wide_characteristic_words}a) we show the 15 words with the highest $\log P \left( w|C \right)$, and in Fig. \ref{fig:total_logp_alegrias} we show the complete $\log P \left( w|C \right)$ distribution for \textit{alegrías}. The essential words for \textit{alegrías} in descending order of $P\left( w | C \right)$ are the following:

\textit{Cádiz} `Cadiz', \textit{llevar} `carry', \textit{Cuando} `When', \textit{dirá} `will say', \textit{conmigo} `with me', \textit{da} `gives', \textit{mandilón} `submissive, pussy-whipped', \textit{veo} `I see', \textit{llaman} `they call', \textit{Muralla Real} `Royal Wall', \textit{vayas} `you go', 'sea', \textit{voy} `I go', \textit{mar}, \textit{adónde} `where to', \textit{vengas} `you come', \textit{más} `more', \textit{quiero} `I want', \textit{dar} `to give', \textit{Que/Qué} `Which/What', \textit{bonita} `beautiful (fem.)', \textit{contigo} `with you', \textit{nombre} `name', \textit{darte} `to give you', \textit{vas} `you go', \textit{niña} `girl', \textit{pasando} `happening', \textit{paseíto} `little walk', \textit{vueltecita} `little turn', \textit{tacón} `heel', \textit{pena} `pain', \textit{patrona} `patroness', \textit{saber} `to know', \textit{sale} `it goes out', \textit{Zaragoza} `Zaragoza', \textit{alegria} `joy', \textit{doble} `double', \textit{relicario} `reliquary', \textit{entender} `to understand', \textit{Navarra} `Navarra', \textit{taconear} `to tap (with heels)', \textit{bota} `boot', \textit{Rosario} `Rosary', \textit{arena} `sand', \textit{San Fernando} `Saint Fernando', \textit{mientan} `they lie', \textit{buenos} `good (masc. pl.)', \textit{barrio} `neighborhood', \textit{tiro} `throw', \textit{están} `they are', \textit{duro} `hard', \textit{madre} `mother', \textit{entrada} `entrance', \textit{valgo} `I am worth', \textit{señora} `lady', \textit{viva} `long life to', \textit{muero} `I die', \textit{titirimundis} `puppets', \textit{dos} `two', \textit{barquero} `boatman', \textit{confianza} `trust', \textit{mujeres} `women', \textit{alma} `soul', \textit{camino} `road', \textit{plata} `silver', \textit{nadie} `nobody', \textit{Una} `One (fem.)', \textit{pago} `I pay', \textit{verte} `to see you', \textit{cayó} `it fell', \textit{prima} `cousin (fem.)', \textit{Yo} `I', \textit{Me} `Me'.

\subsection{\textit{Bulerías}}

In Fig. \ref{fig:15_logprob_bulerias} we show the 15 words with the highest $\log P \left( w|C \right)$, and in Fig. \ref{fig:total_logp_bulerias} we show the complete $\log P \left( w|C \right)$ distribution for \textit{bulerías}. The essential words for \textit{bulerías} in descending order of $P\left( w | C \right)$ are the following: 

\begin{figure*}[t]
    \centering
    \begin{subfigure}[b]{0.5\linewidth}
      \centering
      \includegraphics[width=1\linewidth, angle=0]{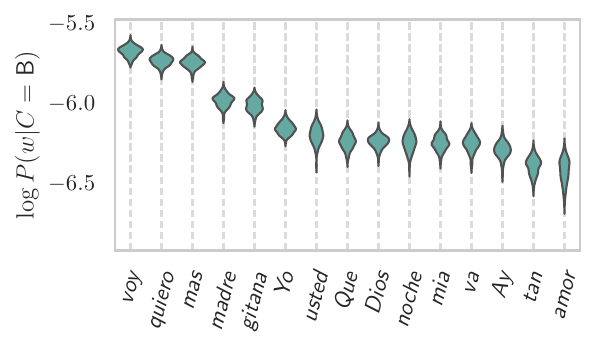}
      \caption{}
      \label{fig:15_logprob_bulerias}
    \end{subfigure}
    \hfill
    \begin{subfigure}[b]{0.4\linewidth}
      \centering
      \includegraphics[width=1\linewidth, angle=0]{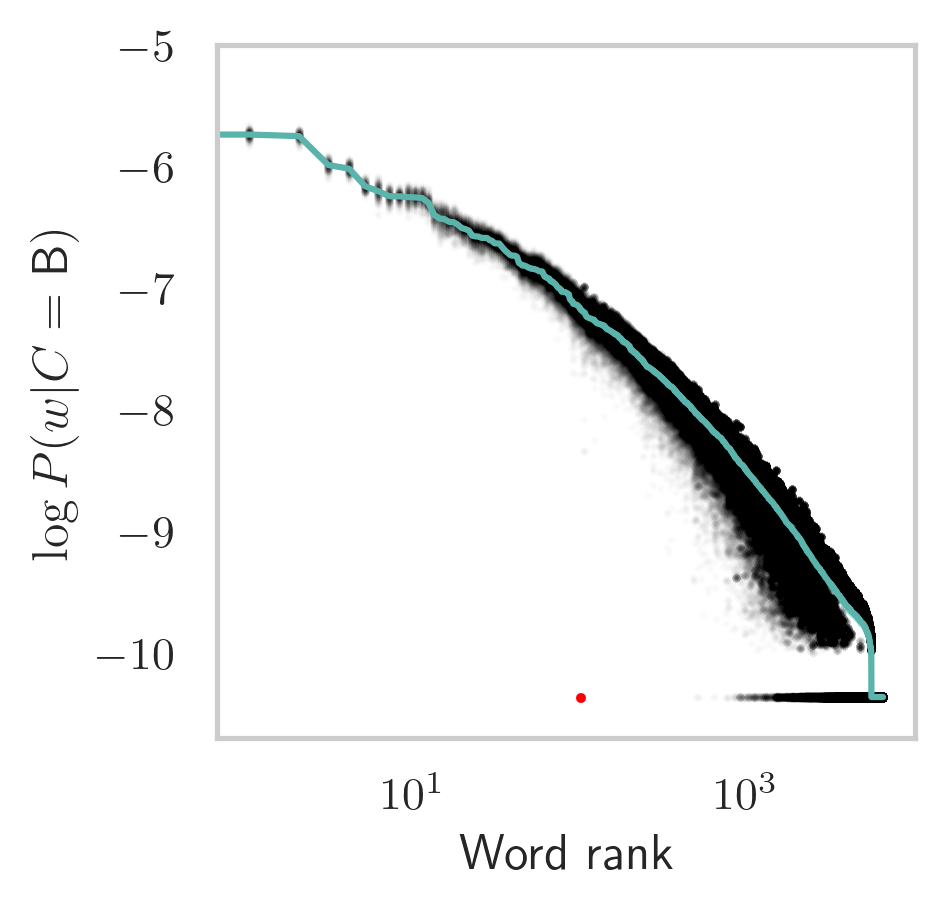}
      \caption{}
      \label{fig:total_logp_bulerias}
    \end{subfigure}
    \caption{(\textbf{a}) The 15 words with highest $\log P \left(w|C \right)$ for \textit{bulerías}, and (\textbf{b}) complete $\log P \left( w|C \right)$ distribution for \textit{bulerías} over 100 trainings. We mark the threshold used to calculate the number of essential words with a red dot (refer to Appendix~\ref{app:essential}).}
    \label{fig:app_bul}
\end{figure*}

\textit{voy} `I go', \textit{quiero} `I want/love', \textit{más} `more', \textit{madre} `mother', \textit{gitana} `gypsy (fem. sing.)', \textit{Yo} `I', \textit{usted} `you (formal)', \textit{Que/Qué} `Which/What', \textit{Dios} `God', \textit{noche} `night', \textit{mía} `mine (fem. sing.)', \textit{va} `goes', \textit{Ay}, \textit{tan} `so (comp.)', \textit{amor} `love', \textit{vino} `wine', \textit{van} `they go', \textit{nadie} `nobody', \textit{vida} `life', \textit{cara} `face', \textit{alma} `soul', \textit{ahora} `now', \textit{contigo} `with you', \textit{gente} `people', \textit{querer} `to love', \textit{Esta} `This (fem.)', \textit{gitano} `gypsy (masc. sing.)', \textit{ver} `to see', \textit{niña} `girl', \textit{mío} `mine', \textit{quiera} `he/she wants', \textit{dos} `two', \textit{calle} `street', \textit{agua} `water', \textit{pena} `pain, sorrow', \textit{Por} `For', \textit{dinero} `money', \textit{corazón} `heart', \textit{pelo} `hair', \textit{quieres} `you want/love', \textit{día} `day', \textit{conmigo} `with me', \textit{vaya} `he/she goes', \textit{Tu} `You', \textit{quiere} `he/she wants', \textit{ser} `to be', \textit{vas} `you go', \textit{casa} `house', \textit{tenía} `he/she had', \textit{aunque} `although', \textit{lleva} `he/she wears', \textit{llevo} `I wear', \textit{mal} `bad', \textit{río} `river', \textit{padre} `father', \textit{digo} `I say', \textit{puedo} `I can', \textit{perdido} `lost', \textit{ojos} `eyes', \textit{tres} `three', \textit{será} `it will be', \textit{No} `No', \textit{mañana} `morning', \textit{mama} `mom', \textit{caballo} `horse', \textit{Cuando} `When', \textit{vera} `side, proximity', \textit{viene} `he/she comes', \textit{grande} `big', \textit{bonita} `beautiful (fem. sing.)', \textit{flores} `flowers', \textit{Él} `He', \textit{hacer} `to do', \textit{cuatro} `four', \textit{quisiera} `I would like', \textit{Virgen} `Virgin', \textit{mundo} `world', \textit{alegría} `joy', \textit{Me} `Me', \textit{niño} `child (masc.)', \textit{mala} `bad', \textit{aquél} `that one', \textit{puerta} `door', \textit{comer} `to eat', \textit{prima} `cousin (fem.)', \textit{campanas} `bells', \textit{cuerpo} `body', \textit{camino} `path', \textit{paso} `step', \textit{dicen} `they say', \textit{loca} `crazy', \textit{Sevilla} `Seville', \textit{María} `María', \textit{Amparo} `Amparo', \textit{dice} `he/she says', \textit{mano} `hand', \textit{daba} `he/she gave', \textit{primo} `cousin (masc.)'.

\subsection{\textit{Fandangos}}

In Fig. \ref{fig:15_logprob_fandangos} we show the 15 words with the highest $\log P \left( w|C \right)$, and in Fig. \ref{fig:total_logp_fandangos} we show the complete $\log P \left( w|C \right)$ distribution for \textit{fandangos}. The essential words for \textit{fandangos} in descending order of $\log P\left( w | C \right)$ in descending order of $P\left( w | C \right)$ are the following:

\begin{figure*}[t]
    \centering
    \begin{subfigure}[b]{0.5\linewidth}
      \centering
      \includegraphics[width=1\linewidth, angle=0]{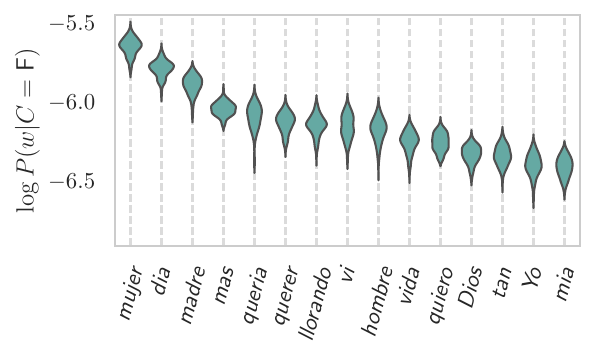}
      \caption{}
      \label{fig:15_logprob_fandangos}
    \end{subfigure}
    \hfill
    \begin{subfigure}[b]{0.4\linewidth}
      \centering
      \includegraphics[width=1\linewidth, angle=0]{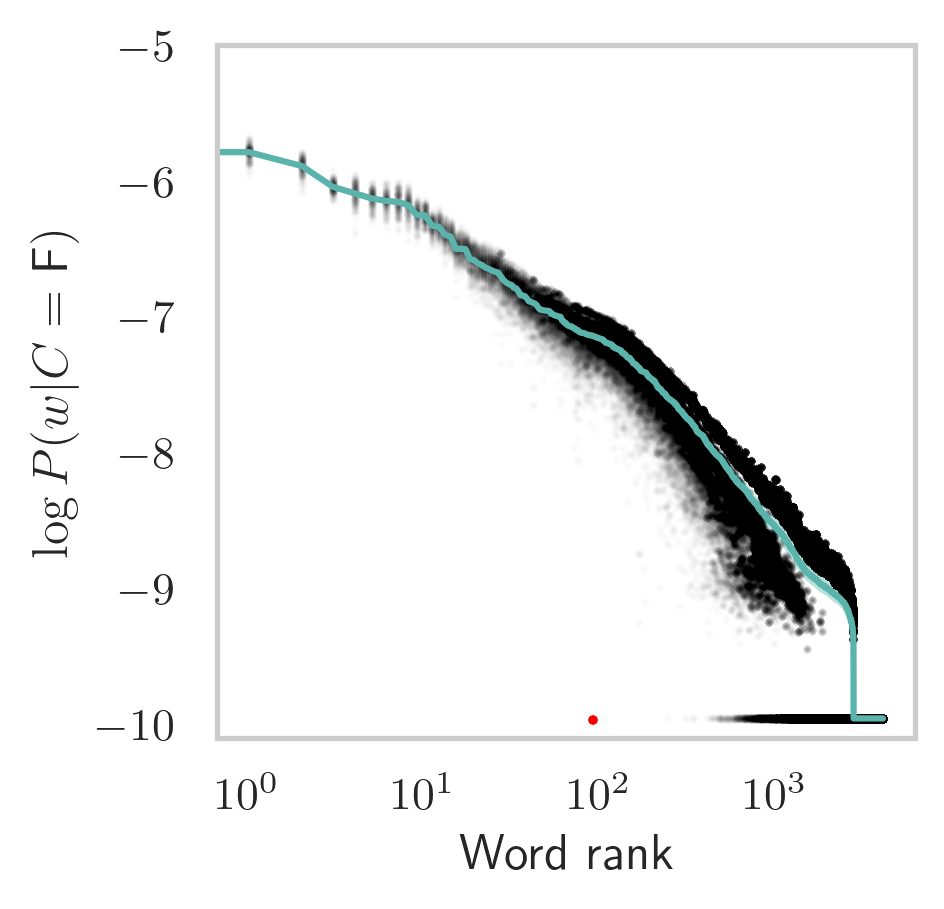}
      \caption{}
      \label{fig:total_logp_fandangos}
    \end{subfigure}
    \caption{(\textbf{a}) The 15 words with highest $\log P \left(w|C \right)$ for \textit{fandangos}, and (\textbf{b}) complete $\log P \left( w|C \right)$ distribution for \textit{fandangos} over 100 trainings. We mark the threshold used to calculate the number of essential words with a red dot (refer to Appendix~\ref{app:essential}).}
    \label{fig:app_fandangos}
\end{figure*}

\textit{mujer} `woman', \textit{día} `day', \textit{madre} `mother', \textit{más} `more', \textit{quería} `I wanted/I loved', \textit{querer} `to want/love', \textit{llorando} `crying', \textit{vi} `I saw', \textit{hombre} `man', \textit{vida} `life', \textit{quiero} `I want/love', \textit{Dios} `God', \textit{tan} `so (comp.)', \textit{Yo} `I', \textit{mía} `mine (fem. sing.)', \textit{voy} `I go', \textit{mundo} `world', \textit{pena} `pain, sorrow', \textit{noche} `night', \textit{ser} `to be', \textit{nadie} `nobody', \textit{quieres} `you want/love', \textit{llorar} `to cry', \textit{De} `Of', \textit{había} `there was', \textit{ver} `to see', \textit{vivir} `to live', \textit{cruz} `cross', \textit{cara} `face', \textit{alma} `soul', \textit{solo} `alone', \textit{dije} `I said', \textit{también} `also', \textit{corazón} `heart', \textit{tenía} `he/she had', \textit{flores} `flowers', \textit{tener} `to have', \textit{agua} `water', \textit{Él} `He', \textit{nunca} `never', \textit{contigo} `with you', \textit{llorado} `cried', \textit{mira} `look', \textit{quiere} `he/she wants', \textit{cinco} `five', \textit{querido} `dear, loved', \textit{vez} `time', \textit{mío} `mine', \textit{loco} `crazy', \textit{penas} `pains, sorrows', \textit{ahora} `now', \textit{pecho} `chest', \textit{dos} `two', \textit{pensamiento} `thought', \textit{paró} `stopped', \textit{traición} `betrayal', \textit{llegar} `to arrive', \textit{boca} `mouth', \textit{gitana} `gypsy (fem.)', \textit{Con} `With', \textit{quiera} `he/she wants', \textit{muerta} `dead (fem. sing.)', \textit{verdad} `truth', \textit{niños} `little boys', \textit{puedo} `I can', \textit{puerta} `door', \textit{Que/Qué} `Which/What', \textit{morena} `brunette', \textit{No} `No', \textit{dinero} `money', \textit{lloran} `they cry', \textit{vera} `side, proximity', \textit{trigal} `wheat field', \textit{sangre} `blood', \textit{Vela} `Candle', \textit{pienso} `I think', \textit{muere} `he/she dies', \textit{acababa} `it was ending', \textit{ninguna} `none (fem.)', \textit{morir} `to die', \textit{días} `days', \textit{decía} `he/she said', \textit{Porque} `Because', \textit{va} `he/she goes', \textit{tarde} `late', \textit{puedes} `you can', \textit{clara} `clear (fem.)', \textit{manos} `hands', \textit{contado} `counted', \textit{sentidos} `senses', \textit{cariño} `affection', \textit{quieras} `you want', \textit{cambio} `change', \textit{gente} `people'.

\subsection{\textit{Malagueñas}} 

In Fig. \ref{fig:15_logprob_malagueñas} we show the 15 words with the highest $\log P \left( w|C \right)$, and in Fig. \ref{fig:total_logp_malagueñas} we show the complete $\log P \left( w|C \right)$ distribution for \textit{malagueñas}. The essential words for \textit{malagueñas} in descending order of $P\left( w | C \right)$ are the following:

\begin{figure*}[t]
    \centering
    \begin{subfigure}[b]{0.5\linewidth}
      \centering
      \includegraphics[width=1\linewidth, angle=0]{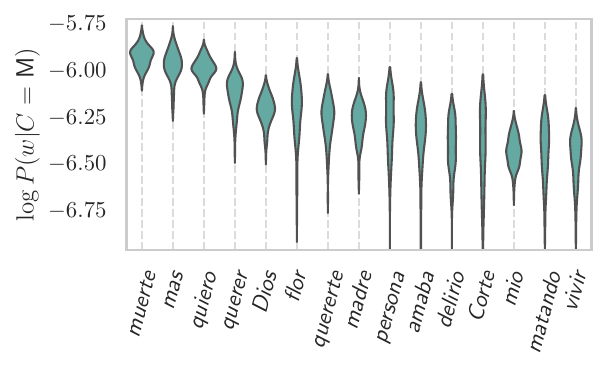}
      \caption{}
      \label{fig:15_logprob_malagueñas}
    \end{subfigure}
    \hfill
    \begin{subfigure}[b]{0.4\linewidth}
      \centering
      \includegraphics[width=1\linewidth, angle=0]{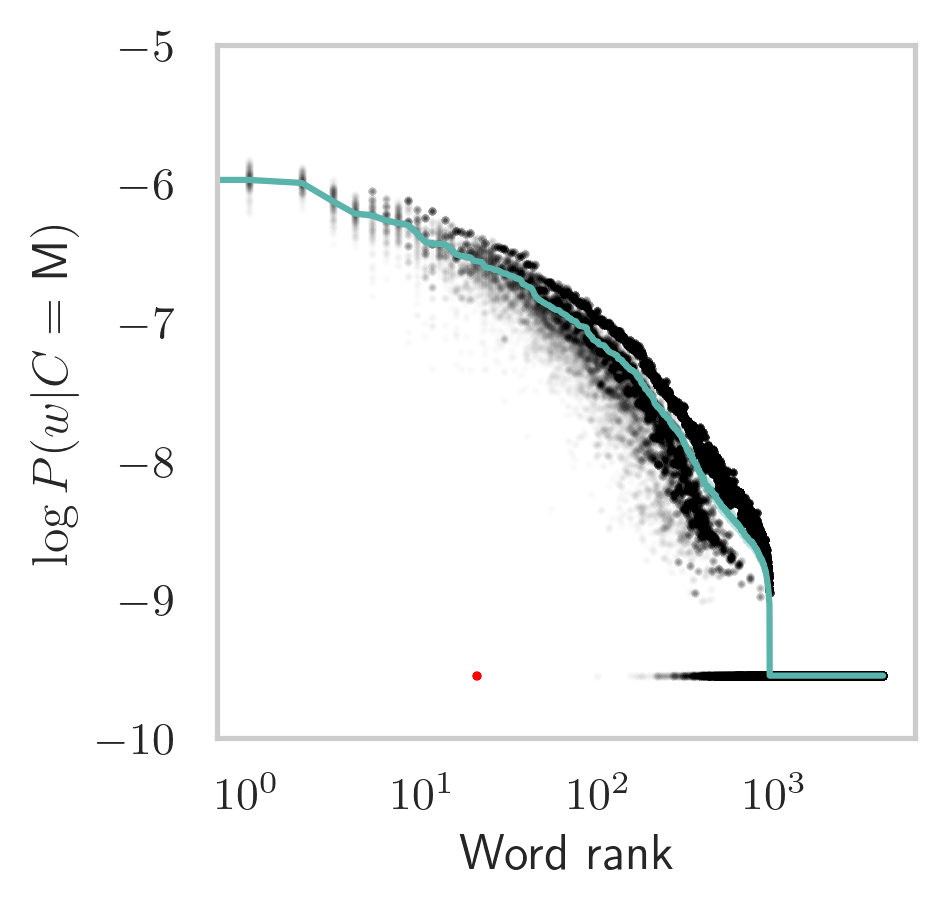}
      \caption{}
      \label{fig:total_logp_malagueñas}
    \end{subfigure}
    \caption{(\textbf{a}) The 15 words with highest $\log P \left(w|C \right)$ for \textit{malagueñas}, and (\textbf{b}) complete $\log P \left( w|C \right)$ distribution for \textit{malagueñas} over 100 trainings. We mark the threshold used to calculate the number of essential words with a red dot (refer to Appendix~\ref{app:essential}).}
    \label{fig:app_malagueñas}
\end{figure*}

\textit{muerte} `death', \textit{más} `more', \textit{quiero} `I want/love', \textit{querer} `to love', \textit{Dios} `God', \textit{flor} `flower', \textit{quererte} `to love you', \textit{madre} `mother', \textit{persona} `person', \textit{amaba} `I loved', \textit{delirio} `delirium', \textit{Corte} `Court', \textit{mío} `mine', \textit{matando} `killing', \textit{vivir} `to live'.

\subsection{\textit{Seguiriyas}}

In Fig. \ref{fig:wide_characteristic_words}c) we show the 15 words with the highest $\log P \left( w|C \right)$, and in Fig. \ref{fig:total_logp_seguiriyas} we show the complete $\log P \left( w|C \right)$ distribution for \textit{seguiriyas}. The essential words for \textit{seguiriyas} in descending order of $P\left( w | C \right)$ are the following: 

\begin{figure}[t]
  \centering
  \includegraphics[width=0.8\linewidth, angle=0]{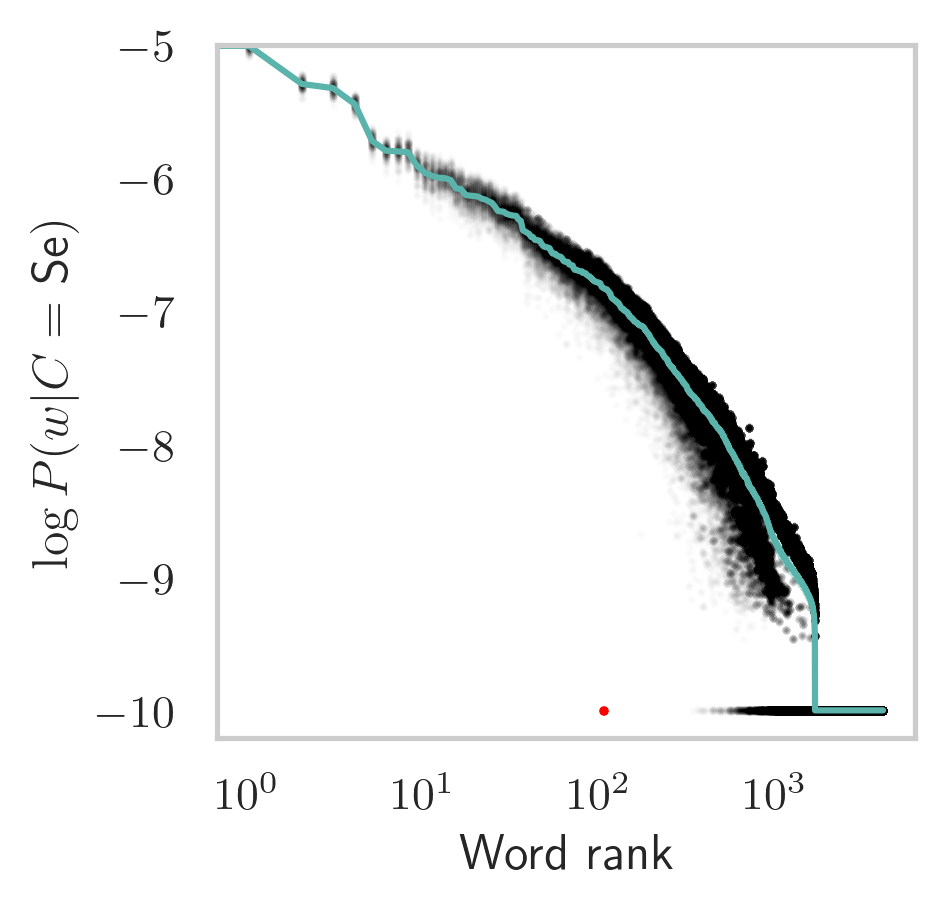}
  \caption{Complete $\log P \left( w|C \right)$ distribution for \textit{seguiriyas} over 100 trainings. We mark the threshold used to calculate the number of essential words with a red dot (refer to Appendix~\ref{app:essential}).}
  \label{fig:total_logp_seguiriyas}
\end{figure}

\textit{madre} `mother', \textit{Dios} `God', \textit{corazón} `heart', \textit{alma} `soul', \textit{más} `more', \textit{mía} `mine (fem. sing.)', \textit{Que/Qué} `Which/What', \textit{mal} `harm', \textit{grandes} `big (pl.)', \textit{mío} `mine', \textit{libertad} `freedom', \textit{andar} `to walk', \textit{encuentro} `I find',  \textit{doy} `I give', \textit{desgracia} `misfortune', \textit{Santiago} `Santiago', \textit{hospital} `hospital', \textit{calor} `heat', \textit{veo} `I see', \textit{rincones} `corners', \textit{pido} `I ask', \textit{llorando} `crying', \textit{fatigas} `fatigues', \textit{quiero} `I want/love', \textit{muero} `I die', \textit{muerte} `death', \textit{voy} `I go', \textit{pago} `I pay', \textit{Siempre} `Always', \textit{Por} `For', \textit{paso} `step', \textit{aliviara} `he/she/I would alleviate', \textit{penas} `pains, sorrows', \textit{duquelas} `pains, sorrows (gipsy lang.)', \textit{noche} `night', \textit{padre} `father', \textit{vida} `life', \textit{reniego} `I renounce', \textit{niños} `children', \textit{cuerpo} `body', \textit{compañerita} `little companion (fem.)', \textit{día} `day', \textit{clavito} `little clove', \textit{adelante} `forward', \textit{clavo} `clove', \textit{pena} `pain, sorrow', \textit{compañera} `companion (fem.)', \textit{dolores} `pains', \textit{puerta} `door', \textit{hecho} `done', \textit{Santa Ana} `Saint Anne', \textit{dos} `two', \textit{nadie} `nobody', \textit{rodillas} `knees', \textit{atrás} `behind', \textit{apelación} `appeal', \textit{rogue} `I beg', \textit{viene} `he/she comes', \textit{Como} `Like, I eat', \textit{grande} `big', \textit{muera} `I die', \textit{caben} `they fit', \textit{Si} `If', \textit{llamarme} `to call me', \textit{sabe} `he/she knows', \textit{lengua} `tongue', \textit{huele} `it smells', \textit{dolor} `pain', \textit{voces} `voices', \textit{No} `No', \textit{pierda} `I lose', \textit{solo} `alone', \textit{perdón} `forgiveness', \textit{vera} `side, proximity', \textit{llamo} `I call', \textit{Yo} `I', \textit{muerto} `dead', \textit{canela} `cinnamon', \textit{daba} `he/she gave', \textit{dieron} `they gave', \textit{encargo} `order', \textit{tan} `so', \textit{Él} `He', \textit{doce} `twelve', \textit{saber} `to know', \textit{llevo} `I carry', \textit{supiera} `he/she knew', \textit{hueles} `you smell', \textit{mando} `I send, I command', \textit{cielo} `sky', \textit{pague} `I/he/she pay', \textit{bien} `well', \textit{mala} `bad (fem.)', \textit{limosna} `alms', \textit{malita} `poor thing (fem.)', \textit{tierra} `land', \textit{vienes} `you come', \textit{sueño} `dream', \textit{médico} `doctor', \textit{horita} `little hour', \textit{dormido} `asleep', \textit{camelo} `I want/love (gipsy lang.)', \textit{eternidad} `eternity', \textit{aquella} `that (fem.)', \textit{hora} `hour', \textit{pasos} `steps', \textit{Te} `You', \textit{había} `there was', \textit{dobles} `double (pl.)', \textit{llores} `you cry', \textit{echo} `I throw', \textit{distinguir} `to distinguish', \textit{San} `Saint', \textit{terelo} `I have', \textit{morir} `to die', \textit{van} `they go', \textit{ir} `to go', \textit{siempre} `always', \textit{dile} `tell him/her', \textit{quemando} `burning', \textit{ver} `to see', \textit{dar} `to give', \textit{mujer} `woman', \textit{Me} `Me', \textit{medio} `half', \textit{siete} `seven', \textit{pasitos} `little steps', \textit{doctor} `doctor', \textit{metida} `inserted (fem. sing.)', \textit{llorar} `to cry', \textit{quieres} `you want', \textit{ladrón} `thief', \textit{remedio} `remedy', \textit{manos} `hands', \textit{señaladitos} `pointed', \textit{ahora} `now', \textit{vine} `I came', \textit{mandé} `I sent', \textit{gente} `people', \textit{solito} `alone', \textit{De} `Of', \textit{calle} `street', \textit{Cuco} `Cuco', \textit{sangre} `blood', \textit{duermes} `you sleep', \textit{vas} `you go', \textit{respondía} `he/she answered', \textit{rodar} `to roll', \textit{negrito} `little black one', \textit{pidiendo} `asking', \textit{días} `days', \textit{malito} `little sick one', \textit{acabo} `I finish', \textit{Eran} `they were', \textit{culpa} `fault', \textit{hermano} `brother', \textit{fatiguitas} `little fatigues', \textit{pasando} `passing', \textit{mundo} `world', \textit{conozco} `I know', \textit{conocí} `I knew', \textit{basta} `enough', \textit{hija} `daughter', \textit{sobra} `it is enough', \textit{Contemplarme} `To contemplate me', \textit{ojitos} `little eyes', \textit{llamar} `to call', \textit{trenzas} `braids', \textit{causas} `causes', \textit{así} `like this', \textit{gitana} `gypsy (fem. sing.)', \textit{muriendo} `dying', \textit{carta} `letter'.

\subsection{\textit{Soleá}}

In Fig. \ref{fig:15_logprob_solea} we show the 15 words with the highest $\log P \left( w|C \right)$, and in Fig. \ref{fig:total_logp_solea} we show the complete $\log P \left( w|C \right)$ distribution for \textit{soleá}. The essential words for \textit{soleá} in descending order of $P\left( w | C \right)$ are the following: 

\begin{figure*}[t]
    \centering
    \begin{subfigure}[b]{0.5\linewidth}
      \centering
      \includegraphics[width=1\linewidth, angle=0]{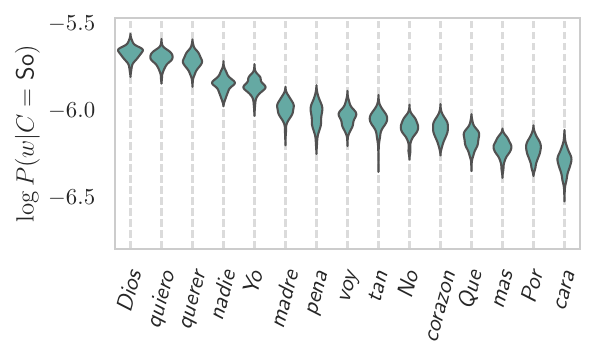}
      \caption{}
      \label{fig:15_logprob_solea}
    \end{subfigure}
    \hfill
    \begin{subfigure}[b]{0.4\linewidth}
      \centering
      \includegraphics[width=1\linewidth, angle=0]{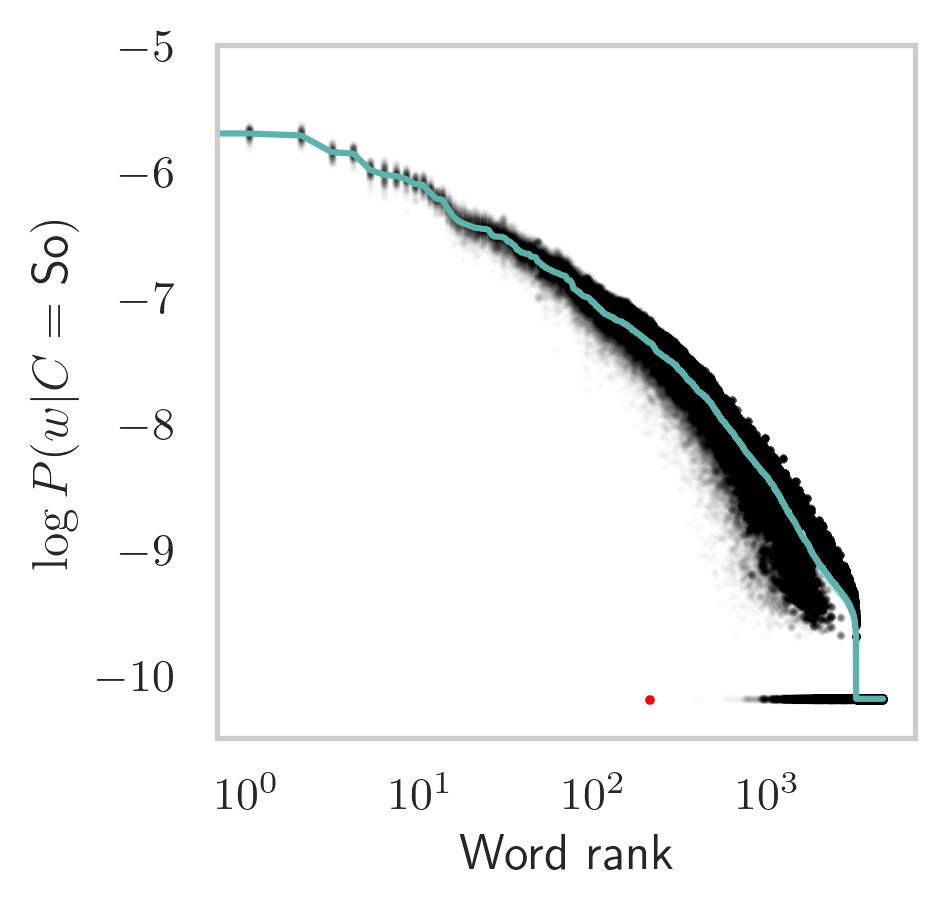}
      \caption{}
      \label{fig:total_logp_solea}
    \end{subfigure}
    \caption{(\textbf{a}) The 15 words with highest $\log P \left(w|C \right)$ for \textit{soleá}, and (\textbf{b}) complete $\log P \left( w|C \right)$ distribution for \textit{soleá} over 100 trainings. We mark the threshold used to calculate the number of essential words with a red dot (refer to Appendix~\ref{app:essential}).}
    \label{fig:app_solea}
\end{figure*}

\textit{Dios} `God', \textit{quiero} `I want/love', \textit{querer} `to love', \textit{nadie} `nobody', \textit{Yo} `I', \textit{madre} `mother', \textit{pena} `pain, sorrow', \textit{voy} `I go', \textit{tan} `so', \textit{No} `No', \textit{corazón} `heart', \textit{Que/Qué} `Which/What', \textit{más} `more', \textit{Por} `For', \textit{cara} `face', \textit{quiera} `I/he/she wants', \textit{mundo} `world', \textit{Undebel} `the devil (gipsy lang.)', \textit{mía} `mine (fem. sing.)', \textit{mal} `bad', \textit{ahora} `now', \textit{ver} `to see', \textit{había} `there was', \textit{gitana} `gypsy (fem.)', \textit{hago} `I do', \textit{dos} `two', \textit{lengua} `tongue', \textit{contigo} `with you', \textit{alma} `soul', \textit{cuerpo} `body', \textit{sol} `sun', \textit{Tu} `Your', \textit{gusto} `taste', \textit{puerta} `door', \textit{puedo} `I can', \textit{quieres} `you want', \textit{vera} `side, proximity', \textit{Si} `If', \textit{tierra} `land', \textit{mala} `bad', \textit{quería} `I/he/she wanted', \textit{Me} `Me', \textit{ser} `to be', \textit{vez} `time', \textit{vas} `you go', \textit{muerte} `death', \textit{mío} `mine', \textit{centro} `center', \textit{nunca} `never', \textit{ojos} `eyes', \textit{metida} `inserted (fem. sing.)', \textit{aire} `air', \textit{penas} `pains, sorrows', \textit{presente} `present', \textit{tiempo} `time', \textit{gente} `people', \textit{prima} `cousin (fem.)', \textit{conmigo} `with me', \textit{calle} `street', \textit{hecho} `done', \textit{hablar} `to talk', \textit{casa} `house', \textit{carnes} `meats', \textit{piedras} `stones', \textit{va} `goes', \textit{vida} `life', \textit{camelo} `I want/love (gipsy lang.)', \textit{día} `day', \textit{creo} `I believe', \textit{sentido} `sense', \textit{bien} `well', \textit{encuentro} `I find, encounter', \textit{compañera} `companion (fem.)', \textit{fatigas} `fatigues', \textit{loco} `crazy', \textit{perdido} `lost', \textit{paso} `step', \textit{tenía} `I/he/she had', \textit{perdone} `I/he/she forgive/s', \textit{sale} `he/she leaves', \textit{verás} `you will see', \textit{flamenca} `flamenca', \textit{viene} `he/she comes', \textit{grande} `big', \textit{quisiera} `I would like/want', \textit{culpa} `fault', \textit{piedra} `stone', \textit{salud} `health', \textit{Él} `He', \textit{echaría} `he/she would throw', \textit{remedio} `remedy', \textit{aunque} `although', \textit{Cuando} `When', \textit{querido} `dear', \textit{loca} `crazy (fem. sing.)', \textit{ciencia} `science', \textit{voces} `voices', \textit{alegría} `joy', \textit{cielo} `sky', \textit{tenerte} `to have you', \textit{ojitos} `little eyes', \textit{persona} `person', \textit{san} `saint', \textit{conoces} `you know', \textit{hablo} `I speak', \textit{colorado} `colored', \textit{pido} `I ask', \textit{campanas} `bells', \textit{morir} `to die', \textit{Sé} `I know', \textit{da} `gives', \textit{malina} `evil (fem. sing.)', \textit{veo} `I see', \textit{camino} `path', \textit{entonces} `then', \textit{Señor} `Lord', \textit{duquelas} `pains, sorrows (gipsy lang.)', \textit{mira} `he/she looks', \textit{pensar} `to think', \textit{perder} `to lose', \textit{pasa} `it happens', \textit{verdad} `truth', \textit{noche} `night', \textit{tener} `to have', \textit{Como} `Like', \textit{Dices} `You say', \textit{quedo} `I stay', \textit{mañana} `morning', \textit{pasar} `to pass', \textit{Tengo} `I have', \textit{quieras} `you want', \textit{vienes} `you come', \textit{vivir} `to live', \textit{Válgame} `Help me', \textit{lleno} `full', \textit{infierno} `hell', \textit{intención} `intention', \textit{muerto} `dead', \textit{Para} `For', \textit{bajabas} `you descended', \textit{llorar} `to cry', \textit{dado} `given', \textit{delante} `in front of', \textit{perdición} `perdition', \textit{borde} `edge', \textit{será} `it will be', \textit{llamo} `I call', \textit{Te} `You', \textit{abrirme} `to open myself', \textit{tres} `three', \textit{padre} `father', \textit{perdón} `forgiveness', \textit{gloria} `glory', \textit{vayas} `you go', \textit{razón} `reason', \textit{pase} `it happens', \textit{murió} `he/she died', \textit{dejado} `left', \textit{eche} `I throw', \textit{perdí} `I lost', \textit{mar} `sea', \textit{vestir} `to dress', \textit{dije} `I said', \textit{La} `The (fem. sing.)', \textit{señalando} `pointing', \textit{queriendo} `wanting/loving', \textit{cuenta} `account', \textit{pasado} `past', \textit{están} `they are', \textit{puesto} `placed', \textit{válgame} `help me', \textit{Con} `With', \textit{ley} `law', \textit{cabecita} `little head', \textit{negra} `black (fem. sing.)', \textit{compañerita} `little companion (fem. sing.)', \textit{llevo} `I carry', \textit{serrana} `mountain woman', \textit{sangre} `blood', \textit{calles} `streets, you shut up', \textit{solamente} `only', \textit{dejarme} `to leave me', \textit{siempre} `always'.

\subsection{\textit{Tangos}}

In Fig. \ref{fig:15_logprob_tangos} we show the 15 words with the highest $\log P \left( w|C \right)$, and in Fig. \ref{fig:total_logp_tangos} we show the complete $\log P \left( w|C \right)$ distribution for \textit{tangos}. The essential words for \textit{tangos} in descending order of $P\left( w | C \right)$ are the following:

\begin{figure*}[t]
    \centering
    \begin{subfigure}[b]{0.5\linewidth}
      \centering
      \includegraphics[width=1\linewidth, angle=0]{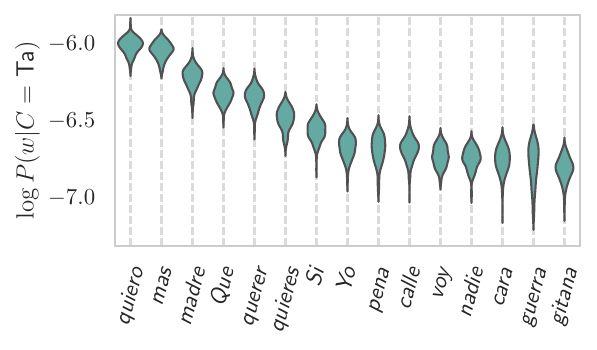}
      \caption{}
      \label{fig:15_logprob_tangos}
    \end{subfigure}
    \hfill
    \begin{subfigure}[b]{0.4\linewidth}
      \centering
      \includegraphics[width=1\linewidth, angle=0]{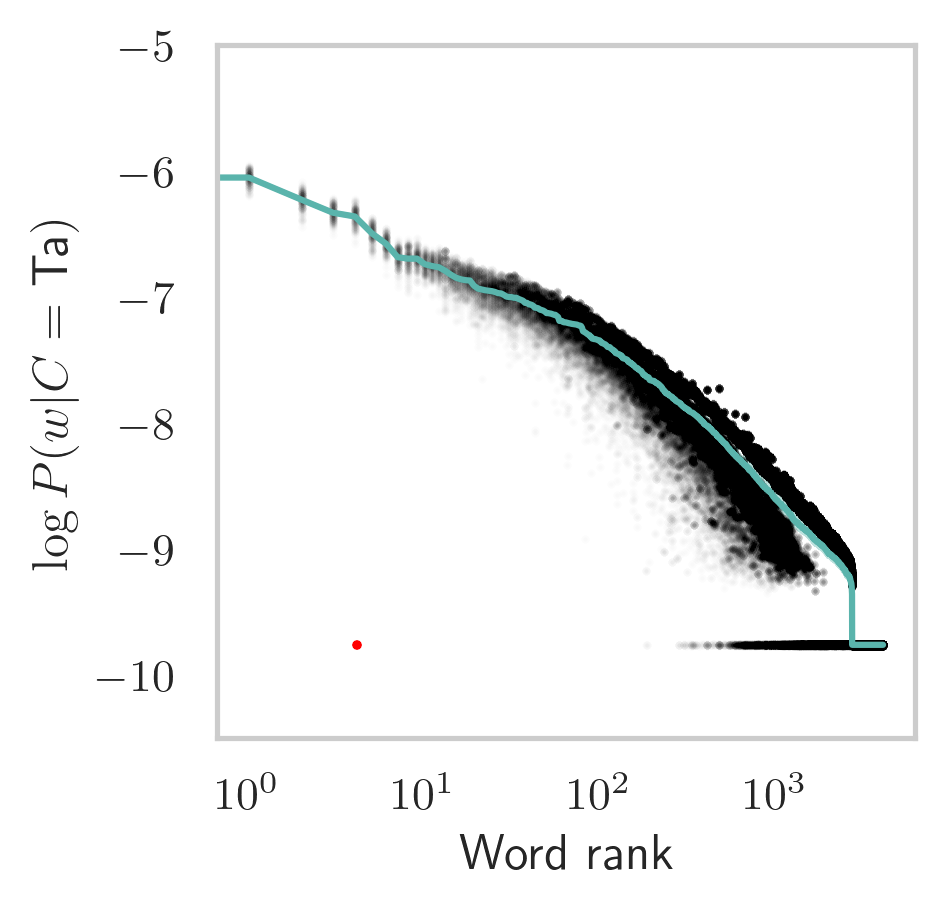}
      \caption{}
      \label{fig:total_logp_tangos}
    \end{subfigure}
    \caption{(\textbf{a}) The 15 words with highest $\log P \left(w|C \right)$ for \textit{tangos}, and (\textbf{b}) complete $\log P \left( w|C \right)$ distribution for \textit{tangos} over 100 trainings. We mark the threshold used to calculate the number of essential words with a red dot (refer to Appendix~\ref{app:essential}).}
    \label{fig:app_tangos}
\end{figure*}

\textit{quiero} `I want/love', \textit{más} `more', \textit{madre} `mother', \textit{Que/Qué} `Which/What', \textit{querer} `to love', \textit{quieres} `you want', \textit{Si} `If', \textit{Yo} `I', \textit{pena} `pain', \textit{calle} `street', \textit{voy} `I go', \textit{nadie} `nobody', \textit{cara} `face', \textit{guerra} `war', \textit{gitana} `gypsy (fem.)', \textit{verás} `you will see', \textit{cuerpo} `body', \textit{Tu} `Your', \textit{vienes} `you come', \textit{Santa María} `Saint Mary', \textit{agua} `water', \textit{mira} `you look, he/she looks', \textit{corazón} `heart', \textit{día} `day', \textit{mía} `mine (fem. sing.)', \textit{noche} `night', \textit{bien} `well', \textit{casa} `house', \textit{mías} `mine', \textit{tonta} `silly', \textit{No} `No', \textit{negros} `black (masc. pl.)', \textit{Dios} `God', \textit{quiera} `he/she wants', \textit{serrana} `mountain woman', \textit{gente} `people', \textit{hace} `he/she does', \textit{alegría} `joy', \textit{vez} `time', \textit{pasa} `it happens', \textit{bonita} `beautiful (fem. sing.)', \textit{mañana} `morning', \textit{Triana} `Triana', \textit{Virgen} `Virgin', \textit{contigo} `with you', \textit{Utrera} `Utrera', \textit{Mi} `My', \textit{mundo} `world', \textit{ser} `to be', \textit{puente} `bridge', \textit{encender} `to light', \textit{niño} `child', \textit{Al} `To the', \textit{fatigas} `fatigues', \textit{flores} `flowers', \textit{Humildad} `Humility', \textit{va} `he/she goes'.

\subsection{\textit{Tientos}}

In Fig. \ref{fig:15_logprob_tientos} we show the 15 words with the highest $\log P \left( w|C \right)$, and in Fig. \ref{fig:total_logp_tientos} we show the complete $\log P \left( w|C \right)$ distribution for \textit{tientos}. The essential words for \textit{tientos} in descending order of $P\left( w | C \right)$ are the following: 

\begin{figure*}[t]
    \centering
    \begin{subfigure}[b]{0.5\linewidth}
      \centering
      \includegraphics[width=1\linewidth, angle=0]{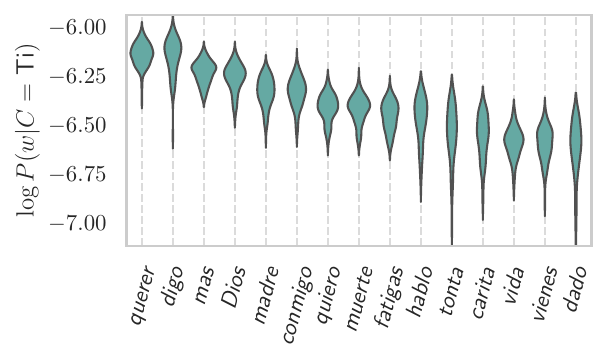}
      \caption{}
      \label{fig:15_logprob_tientos}
    \end{subfigure}
    \hfill
    \begin{subfigure}[b]{0.4\linewidth}
      \centering
      \includegraphics[width=1\linewidth, angle=0]{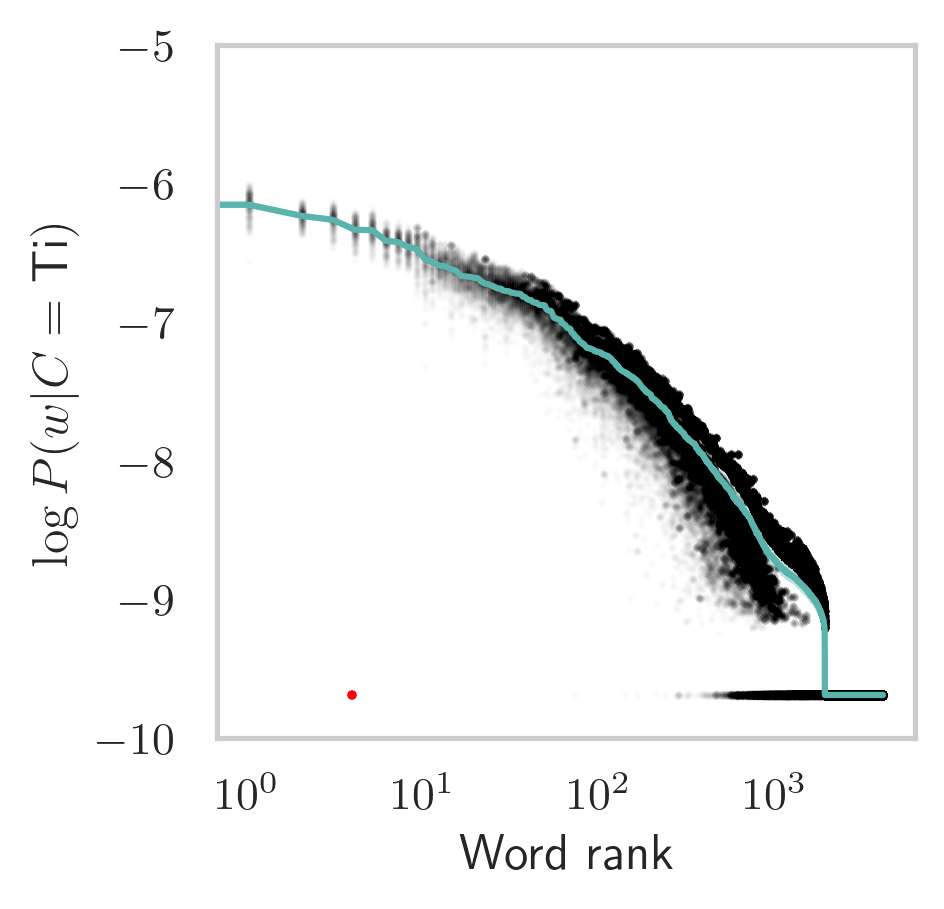}
      \caption{}
      \label{fig:total_logp_tientos}
    \end{subfigure}
    \caption{(\textbf{a}) The 15 words with highest $\log P \left(w|C \right)$ for \textit{tientos}, and (\textbf{b}) complete $\log P \left( w|C \right)$ distribution for \textit{tientos} over 100 trainings. We mark the threshold used to calculate the number of essential words with a red dot (refer to Appendix~\ref{app:essential}).}
    \label{fig:app_tientos}
\end{figure*}

\textit{querer} `to want/love', \textit{digo} `I say', \textit{más} `more', \textit{Dios} `God', \textit{madre} `mother', \textit{conmigo} `with me', \textit{quiero} `I want/love', \textit{muerte} `death', \textit{fatigas} `fatigues', \textit{hablo} `I speak', \textit{tonta} `silly (fem. sing.)', \textit{carita} `little face', \textit{vida} `life', \textit{vienes} `you come', \textit{tan} `so (comp.)', \textit{dado} `given', \textit{llorar} `to cry', \textit{pocito} `little well', \textit{mía} `mine (fem. sing.)', \textit{grande} `big (sing.)', \textit{hecho} `made (masc. sing.)', \textit{cuerpo} `body', \textit{día} `day', \textit{motivo} `reason', \textit{siento} `I feel', \textit{serrana} `mountain woman', \textit{Yo} `I', \textit{pena} `pain, sorrow', \textit{gitana} `gypsy (fem. sing.)', \textit{veo} `I see', \textit{vera} `side, proximity', \textit{inocente} `innocent (sing.)', \textit{corazón} `heart', \textit{Virgen} `Virgin', \textit{daré} `I will give', \textit{voy} `I go', \textit{bien} `well', \textit{agüita} `little water', \textit{aquel} `that one', \textit{rico} `rich (masc. sing.)', \textit{Si} `If', \textit{verita} `little \textit{vera}', \textit{Que/Qué} `Which/What', \textit{temblé} `I trembled', \textit{dan} `they give', \textit{verte} `to see you', \textit{escucharlo} `to listen to it', \textit{agua} `water', \textit{parece} `it seems', \textit{mentira} `lie', \textit{vente} `you come', \textit{contigo} `with you', \textit{grandes} `big (pl.)', \textit{dando} `giving', \textit{guerra} `war'.

\medskip

%% file: app_stopwords.tex
\textit{a}, \textit{ah}, \textit{ay}, \textit{al}, \textit{algo}, \textit{algunas}, \textit{algunos}, \textit{ante}, \textit{antes}, \textit{co}, \textit{como}, \textit{con}, \textit{contra}, \textit{cual}, \textit{cuando}, \textit{de}, \textit{del}, \textit{desde}, \textit{donde}, \textit{durante}, \textit{e}, \textit{el}, \textit{ella}, \textit{ellas}, \textit{ellos}, \textit{en}, \textit{entre}, \textit{era}, \textit{erais}, \textit{eran}, \textit{eras}, \textit{eres}, \textit{es}, \textit{esa}, \textit{esas}, \textit{ese}, \textit{eso}, \textit{esos}, \textit{esta}, \textit{estaba}, \textit{estabais}, \textit{estaban}, \textit{estabas}, \textit{estad}, \textit{estada}, \textit{estadas}, \textit{estado}, \textit{estados}, \textit{estamos}, \textit{estando}, \textit{estar}, \textit{estaremos}, \textit{estas}, \textit{este}, \textit{estemos}, \textit{esto}, \textit{estos}, \textit{estoy}, \textit{estuve}, \textit{estuviera}, \textit{estuvierais}, \textit{estuvieran}, \textit{estuvieras}, \textit{estuvieron}, \textit{estuviese}, \textit{estuvieseis}, \textit{estuviesen}, \textit{estuvieses}, \textit{estuvimos}, \textit{estuviste}, \textit{estuvisteis}, \textit{estuvo}, \textit{fue}, \textit{fuera}, \textit{fuerais}, \textit{fueran}, \textit{fueras}, \textit{fueron}, \textit{fuese}, \textit{fueseis}, \textit{fuesen}, \textit{fueses}, \textit{fui}, \textit{fuimos}, \textit{fuiste}, \textit{fuisteis}, \textit{ha}, \textit{habida}, \textit{habidas}, \textit{habido}, \textit{habidos}, \textit{habiendo}, \textit{habremos}, \textit{han}, \textit{has}, \textit{hasta}, \textit{hay}, \textit{haya}, \textit{hayamos}, \textit{hayan}, \textit{hayas}, \textit{he}, \textit{hemos}, \textit{hube}, \textit{hubiera}, \textit{hubierais}, \textit{hubieran}, \textit{hubieras}, \textit{hubieron}, \textit{hubiese}, \textit{hubieseis}, \textit{hubiesen}, \textit{hubieses}, \textit{hubimos}, \textit{hubiste}, \textit{hubisteis}, \textit{hubo}, \textit{i}, \textit{la}, \textit{las}, \textit{le}, \textit{les}, \textit{lo}, \textit{los}, \textit{me}, \textit{mi}, \textit{mis}, \textit{mucho}, \textit{muchos}, \textit{muy}, \textit{nada}, \textit{ni}, \textit{no}, \textit{nos}, \textit{nosotras}, \textit{nosotros}, \textit{nuestra}, \textit{nuestras}, \textit{nuestro}, \textit{nuestros}, \textit{o}, \textit{oh}, \textit{os}, \textit{otra}, \textit{otras}, \textit{otro}, \textit{otros}, \textit{para}, \textit{pa}, \textit{pero}, \textit{poco}, \textit{por}, \textit{porque}, \textit{que}, \textit{quien}, \textit{quienes}, \textit{se}, \textit{sea}, \textit{seamos}, \textit{sean}, \textit{seas}, \textit{seremos}, \textit{si}, \textit{sido}, \textit{siendo}, \textit{sin}, \textit{sobre}, \textit{sois}, \textit{somos}, \textit{son}, \textit{soy}, \textit{su}, \textit{sus}, \textit{suya}, \textit{suyas}, \textit{suyo}, \textit{suyos}, \textit{tanto}, \textit{te}, \textit{tendremos}, \textit{tened}, \textit{tenemos}, \textit{tenga}, \textit{tengamos}, \textit{tengan}, \textit{tengas}, \textit{tengo}, \textit{tenida}, \textit{tenidas}, \textit{tenido}, \textit{tenidos}, \textit{teniendo}, \textit{ti}, \textit{tiene}, \textit{tienen}, \textit{tienes}, \textit{todo}, \textit{todos}, \textit{tu}, \textit{tus}, \textit{tuve}, \textit{tuviera}, \textit{tuvierais}, \textit{tuvieran}, \textit{tuvieras}, \textit{tuvieron}, \textit{tuviese}, \textit{tuvieseis}, \textit{tuviesen}, \textit{tuvieses}, \textit{tuvimos}, \textit{tuviste}, \textit{tuvisteis}, \textit{tuvo}, \textit{tuya}, \textit{tuyas}, \textit{tuyo}, \textit{tuyos}, \textit{u}, \textit{un}, \textit{una}, \textit{uno}, \textit{unos}, \textit{vosotras}, \textit{vosotros}, \textit{vuestra}, \textit{vuestras}, \textit{vuestro}, \textit{vuestros}, \textit{y}, \textit{ole}, \textit{arsa}, \textit{b}, \textit{c}, \textit{d}, \textit{f}, \textit{g}, \textit{h}, \textit{j}, \textit{k}, \textit{l}, \textit{m}, \textit{n}, \textit{p}, \textit{q}, \textit{r}, \textit{s}, \textit{t}, \textit{v}, \textit{w}, \textit{x}, \textit{y}, \textit{z}.